\documentclass[runningheads]{llncs}

\usepackage{eccv}

\usepackage{eccvabbrv}
\definecolor{ourscore}{HTML}{F1F5FB}
\definecolor{tiffany}{RGB}{10,186,181}

\usepackage{wrapfig}
\usepackage{float} %
\usepackage{mathtools}
\usepackage{amsfonts}

\usepackage{graphicx}
\usepackage{pifont}
\usepackage{threeparttable}
\usepackage{multirow}

\usepackage{amsmath}  
\usepackage{arydshln}
\usepackage{bbding}
\usepackage{booktabs}
\usepackage{url}
\usepackage{bm}

\usepackage{colortbl}

\def\Phi{\boldsymbol{\phi}}

\def\Omegga{\boldsymbol{\omega}}
\def\S{\mathcal{S}}

\def\S{\mathcal{S}}

\definecolor{mygreen}{HTML}{538234}
\definecolor{myred}{HTML}{c00000}
\definecolor{mypurple}{HTML}{7030A0}
\definecolor{myblue}{HTML}{F1F5FB}
\definecolor{mydblue}{HTML}{E3EBF5}
\usepackage{cuted}

\usepackage[linesnumbered,ruled,vlined]{algorithm2e}
\usepackage{multicol}  

\definecolor{cvprblue}{rgb}{0.21,0.49,0.74}
\usepackage{newfloat}
\usepackage{listings}

\usepackage{graphicx}
\usepackage{booktabs}

\usepackage[accsupp]{axessibility}  

\usepackage{hyperref}

\usepackage{orcidlink}

\begin{document}

\title{Learning with Bilevel-Minimax Optimization for Efficient and Reliable  Transfer Attacks} 

\titlerunning{Bilevel-Minimax Adversarial Transfer}

\author{Yaohua Liu\inst{1}\orcidlink{0000-0002-9057-1645} \and
Yifan Guo\inst{2}\orcidlink{0009-0003-9481-6253} \and
Jiaxin Gao\inst{3}\orcidlink{0000-0002-0023-1269}\thanks{Corresponding author.}}

\authorrunning{Y.~Liu et al.}

\institute{The University of Hong Kong, Hong Kong SAR, China\\
\email{liuyaohua.918@gmail.com} \and
International School of Information Science \& Engineering, Dalian University of Technology, Dalian, China\\
\email{friscoguo@gmail.com} \and
The Hong Kong Polytechnic University, Hong Kong SAR, China\\
\email{jiaxinn.gao@outlook.com}}

\maketitle

\begin{abstract}
Transfer-based adversarial attacks craft adversarial examples using surrogate models to mislead black-box victim models.  Beyond perturbation generation, transferability is fundamentally governed by the coupling of initialization, surrogate adaptation, and gradient dynamics. We revisit this challenge from a \texttt{Bilevel-Minimax} perspective and instantiate it in \textcolor{tiffany}{\texttt{\textbf{BMAT}}} (\texttt{\textbf{\textcolor{tiffany}{B}ilevel-\textcolor{tiffany}{M}inimax \textcolor{tiffany}{A}dversarial \textcolor{tiffany}{T}ransfer}}). The bilevel formulation captures the dependency between initialization and perturbation, while the inner minimax problem promotes surrogate robustness for cross-architecture generalization. Algorithmically, we design an integrated bottom-up solver that combines a Soft Weight Modulator and an Implicit Gradient Approximator for ternary coupling interaction.  
We further provide theoretical insights into the optimization dynamics of the proposed bilevel-minimax framework.
Extensive experiments on classification and segmentation benchmarks show that \texttt{BMAT} surpasses \texttt{10+} strong baselines across \texttt{30+} victim models, improving both intra- and cross-architecture transfer, and yielding up to \texttt{2×} mIoU reduction. 
Code is available at \url{https://github.com/callous-youth/BMAT}.
\keywords{Transfer attacks \and Bilevel-Minimax Optimization\and Classification \and Semantic Segmentation}
\end{abstract}

\section{Introduction}
\label{sec:intro}

Adversarial attacks~\cite{gu2022segpgd,bai2025rat,yin2025adversarial,jiao2023pearl,liu2026past} craft imperceptible perturbations to mislead deep neural networks, posing serious threats to real-world vision systems~\cite{rebuffi2022revisiting,carlini2017towards,agnihotri2024cospgd,gao2024dual}. Among these, \emph{transfer-based adversarial attacks} are particularly concerning as they generate Adversarial Examples (AEs) against surrogate models that can transfer to unknown victim models without query access~\cite{dong2018boosting,xie2019improving,wang2024boosting}. This black-box characteristic makes transfer attacks both highly practical and challenging to defend against~\cite{nakka2020indirect,9150896,he2023transferable,li2023making}.

\begin{figure*}[t]
	\centering
	\includegraphics[width=0.99\textwidth]{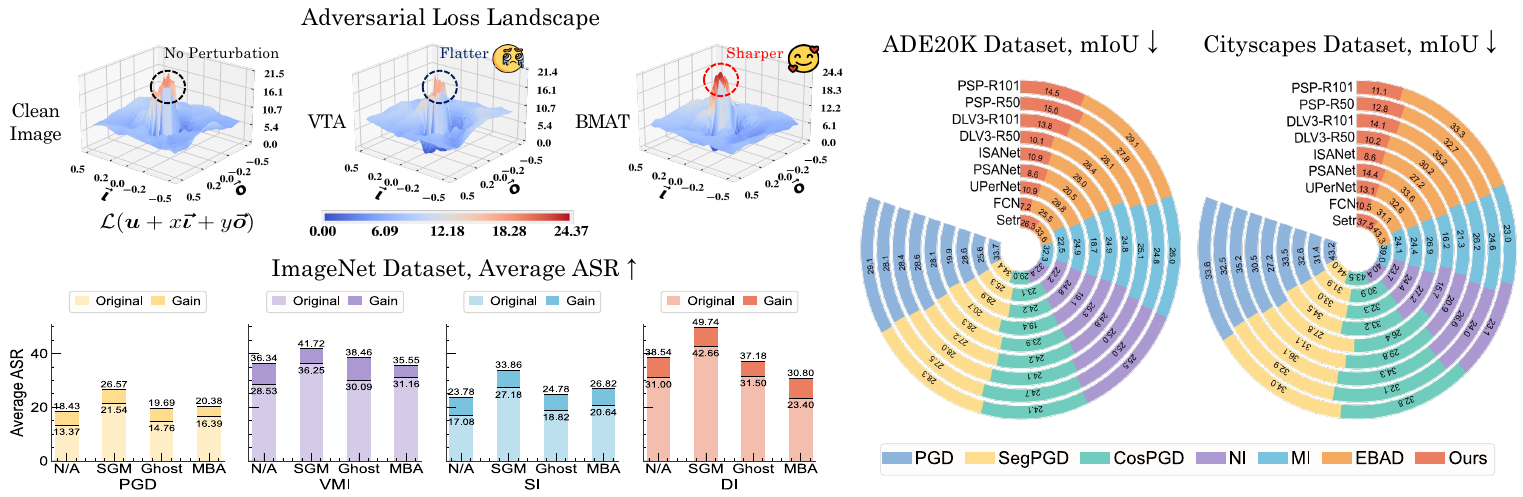}
	\caption{
		\textbf{Landscape (Top Left):} 
		We visualize the loss $\mathcal{L}(\boldsymbol{u} + x{\vec{\iota}}+y{\vec{o}})$ along sign-gradient direction ${\vec{\iota}}$ and random direction ${\vec{o}}$, centered at the original image $\boldsymbol{u}$.
		Compared with VTA, \texttt{BMAT} reaches a \emph{higher loss region} with stronger cross-model transfer behavior.
		\textbf{Classification (Bottom Left):} \texttt{BMAT} achieves a $\mathbf{26.2\%}$ ASR gain on ImageNet~\cite{russakovsky2015imagenet} across 10 victims with 16 base attackers.
		\textbf{Segmentation (Right):} \texttt{BMAT} reduces mIoU by $\mathbf{46.4\%}$ on ADE20K~\cite{zhou2017scene} and $\mathbf{43.7\%}$ on Cityscapes~\cite{cordts2016cityscapes}.
	}
	\label{fig:FirstFigure}  
\end{figure*}
Prior methods have explored transferability from several perspectives, such as surrogate structure modifications~\cite{wang2023rethinking,wang2021feature,papernot2016transferability,huang2019black}, input transformation techniques~\cite{xie2019improving,long2022frequency,liu2024boosting}, and momentum-based strategies~\cite{dong2018boosting,lin2019nesterov}. Some ensemble-based methods~\cite{li2023making,cai2023ensemble} enhance transferability by leveraging ensemble gradients from multiple surrogate models. Other studies~\cite{miller2020query,lord2022attacking,wang2025improving} also explore transferability by employing stronger surrogate models, often at the cost of additional training and task-specific loss designs. Several efforts~\cite{fang2022learning,du2019query} have explored the role of initialization through customized schemes, typically combining data augmentation and surrogate ensembles in a hand-crafted manner.  
However, most approaches still optimize isolated factors such as perturbation design or surrogate structure, while treating other variables as fixed or heuristic. This fragmented treatment fails to capture the interaction dynamics among initialization, perturbation, and surrogate adaptation, limiting both transferability and optimization efficiency.

We posit that transferability fundamentally arises from the \emph{ternary coupling interaction} among: the \textbf{initialization perturbation (IP)}, which seeds the attack trajectory and determines the explored perturbation regions; the \textbf{adversarial perturbation}, which exploits model-specific vulnerabilities; and the \textbf{surrogate parameter}, which shapes the gradient landscape. When these variables are tuned independently, the attack dynamics often degenerate into a standard white-box procedure on the surrogate, overfitting surrogate-specific artifacts rather than promoting cross-model transfer. Existing methods largely adopt factor-wise and heuristic designs, resulting in fragmented optimization dynamics. This decoupled paradigm leads to two key limitations: 
(i) \emph{misaligned optimization dynamics}, where separately tuned variables fail to produce coherent cross-model transfer behavior; and 
(ii) \emph{the absence of a unified optimization formulation} to systematically capture and coordinate these interdependent factors. These observations prompt a fundamental question:
\noindent \textit{How can we develop a unified optimization framework that explicitly models and jointly coordinates these interacting factors to enhance transferability in a principled manner?}

\subsection{Contributions}

To bridge these gaps, we formalize the above hierarchical dependencies in a unified \texttt{Bilevel-Minimax} optimization framework, casting the ternary interactions of transfer attacks into a mathematically principled formulation. \textcolor{tiffany}{\texttt{\textbf{BMAT}}} (\texttt{\textbf{\textcolor{tiffany}{B}ilevel-\textcolor{tiffany}{M}inimax \textcolor{tiffany}{A}dversarial \textcolor{tiffany}{T}ransfer}}) formulates transfer attacks as a bilevel-with-minimax problem: the inner minimax jointly adapts the perturbation and the surrogate’s soft weights (one backward pass) to surface robust, transferable gradients, while the outer level learns an IP using a conjugate-gradient hypergradient that avoids unrolling. 
Algorithmically, we design a tailored bottom-up solver, where {Soft Weight Modulator (SWM)} performs single-step joint updates of perturbations and soft surrogate weights, and Implicit Gradient Approximator (IGA) refines IP using implicit feedback without incurring the overhead of nested gradient computations.  
Theoretically, we analyze the joint optimization dynamics of the regularized \texttt{BMAT} solver, offering insights into its stability and descent behavior.
Fig.~\ref{fig:FirstFigure} shows \texttt{BMAT} consistently boosts AE transferability across attackers in classification and segmentation.
The main contributions are summarized as follows:
\begin{itemize}
	\item \textbf{Formulation.} \texttt{BMAT} casts transfer attacks into a unified \texttt{Bilevel-Minimax} framework that  \emph{explicitly couples} the initialization, perturbation, and surrogate, thereby enabling \emph{hierarchical dynamic coordination}.
	\item \textbf{Algorithm.} Our bottom-up solver orchestrates a  SWM for cross-architecture generalization and an IGA for adaptive initialization guidance, enabling efficient \emph{trajectory seeding} and \emph{warm-started fast transfer}.

	\item \textbf{Analysis.} We provide a stability-oriented analysis of the regularized bilevel-minimax solver, characterizing the joint optimization dynamics of these coupled variables.
	\item \textbf{Experiments.} Evaluations cross \texttt{30+} victim models and  2 tasks (classification and segmentation) show that  \textcolor{black}{\texttt{BMAT}} consistently improves both intra- and cross-architecture transferability. 
\end{itemize}

\begin{figure*}[!t]
	\centering 
	\begin{tabular}{@{\extracolsep{-0.2em}}c@{\extracolsep{0.1em}}}	 
		\includegraphics[width=11.5cm,trim=1 0 0 0, clip]{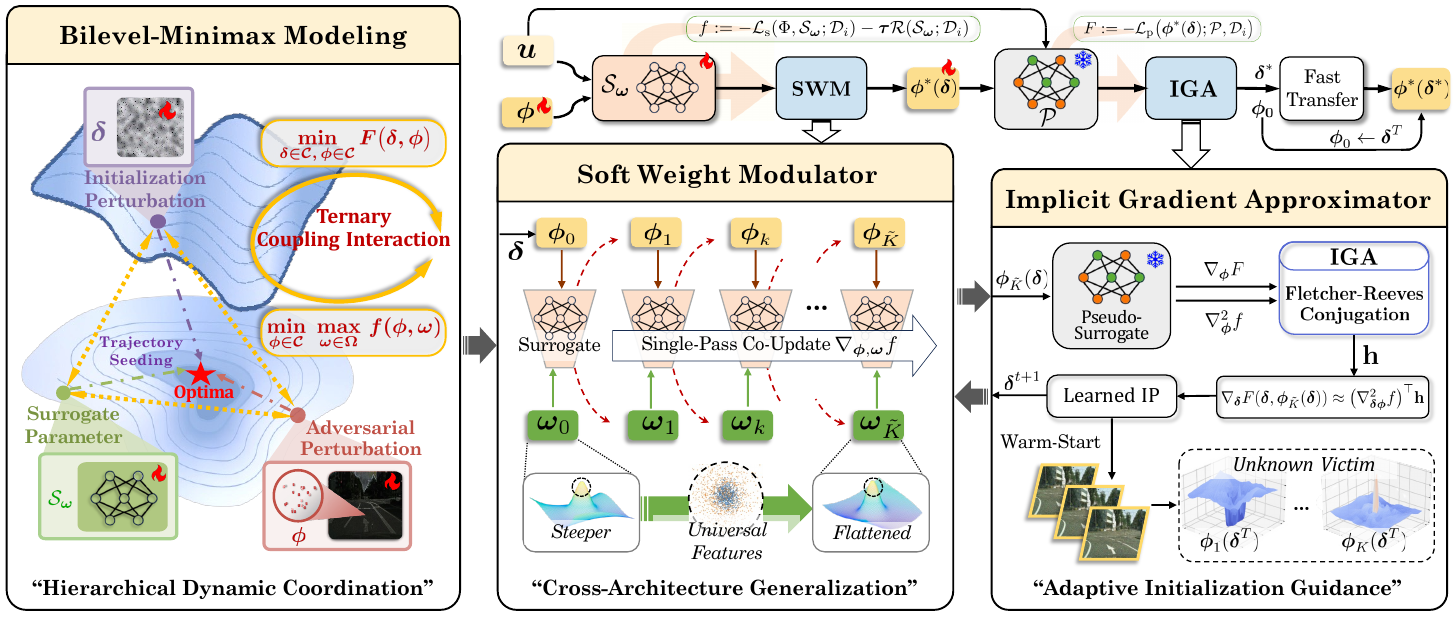}\\   
	\end{tabular}
	\caption{\text{\bf BMAT pipeline. } \textit{Left}: Bilevel-Minimax formulation illustrating the ternary coupling interaction across three variables. \textit{Top}: Overall data flow and supervision in \texttt{BMAT}.  \textit{Middle}: SWM showing inner-loop single-pass co-update of perturbation and soft surrogate weights to obtain cross-architecture gradients. \textit{Right}: IGA depicting hypergradient calculation of IP using Fletcher-Reeves Conjugation. }
	\label{fig:visual_3}
\end{figure*}%

\section{Related Work}

\textbf{Input- and Gradient-based Transfer Attacks.}
Most transfer attacks operate on a fixed surrogate and improve transferability by reshaping gradients and/or transforming the input during iteration~\cite{gu2023survey,ododo2025adversarial,li2025enhancing}. Momentum-based methods such as MI~\cite{dong2018boosting}, NI~\cite{lin2019nesterov}, and VMI~\cite{wang2021enhancing} stabilize trajectories by accumulating or reweighting historical gradients. In parallel, input-transformation techniques, including DI~\cite{xie2019improving}, TI~\cite{dong2019evading}, SI~\cite{lin2019nesterov}, and Admix~\cite{wang2021admix}, inject diversity via random resizing, padding, translation, or patch mixing to reduce overfitting~\cite{yuan2026riemannian}. Objective-level designs further encode architectural or task priors, e.g., SGM~\cite{wu2020boosting} and Ghost~\cite{li2020learning} for classification~\cite{wang2023rethinking,chen2025enhancing,guo2020backpropagating,papernot2016transferability}, and SegPGD~\cite{gu2022segpgd} or CosPGD~\cite{agnihotri2024cospgd} for dense prediction.  Recent work such as RAP~\cite{qin2022boosting} encourages flatter loss landscapes via repeated explicit maximization, but remains within a single-level minimax framework. \emph{Summary.}  Despite their strong empirical performance, these attacks still treat the surrogate and initialization as fixed design choices and optimize only the perturbation in a single-level manner, lacking a principled formulation of why and how perturbations transfer across models.

\noindent\textbf{Surrogate Ensembles and Adaptation.}
A complementary thread manipulates the surrogate side via ensembles or adaptive modeling. Ensemble-based attacks aggregate gradients from multiple architectures to reduce model bias~\cite{li2023making,chen2023rethinking}, and dynamic reweighting further stabilizes multi-model signals~\cite{cai2023ensemble}. Beyond ensembling, GFCS~\cite{lord2022attacking} exploits co-image sampling, undertrained surrogates flatten loss landscapes~\cite{miller2020query}, DRA~\cite{zhu2022toward} regularizes perturbations distributionally, and FAUG~\cite{wang2025improving} augments surrogate features. Meta-learning strategies~\cite{yuan2021meta,fang2022learning} learn cross-model update rules but usually rely on surrogate ensembles or tailored training pipelines. BETAK~\cite{liu2024advancing} introduces ensemble-guided initialization learning, yet still lacks a unified optimization formulation. \emph{Summary.} While these methods emphasize the surrogate as a key component, surrogate adaptation is usually decoupled from perturbation generation and initialization, or achieved at high cost via large ensembles.

\noindent\textbf{Bilevel Optimization in Adversarial Robustness.}
Bilevel optimization~\cite{liu2021investigating,liu2023averaged,liu2021towards,mu2021triple} provides a principled framework for hierarchical learning~\cite{franceschi2018bilevel,ji2021bilevel,gao2024collaborative}. A prominent line reformulates adversarial training as a bilevel problem, where the inner problem minimizes perturbed-data loss and the outer problem updates model parameters for robustness, leading to more stable and efficient training schemes~\cite{zhang2022revisiting,liu2021value,fan2024bi,gao2025learning}. Beyond adversarial training, bilevel formulations have also been used to tune robustness-related hyperparameters~\cite{macdonald2019adversarial} and adaptive attack budgets~\cite{li2020automa}. \emph{Positioning.} In contrast, our work is, to our knowledge, among the first to systematically apply a \texttt{Bilevel-Minimax} formulation to \emph{transfer-based black-box attacks}, explicitly coupling three variables within a single optimization scheme rather than treating them as fixed or independent components.

\section{Methodology}

\subsection{Preliminary}\label{sec:VTA}

We begin by formalizing the standard transfer attack paradigm. Let $\mathcal{D}_i = (\boldsymbol{u}_i, \boldsymbol{v}_i)$ denote the $i$-th data pair from dataset $\mathcal{D} = \{(\boldsymbol{u}_i, \boldsymbol{v}_i)\}_{i=1}^M$, and $\mathcal{S}_{\bm{\omega}}$ represent the surrogate model parameterized by $\bm{\omega} \in \Omega$. The surrogate loss on $\mathcal{D}_i$ is written as $\mathcal{L}_{\text{s}}(\mathcal{S}_{\bm{\omega}}(\boldsymbol{u}_i), \boldsymbol{v}_i)$.
In Vanilla Transfer Attack (VTA)~\cite{kurakin2018adversarial}, the objective is to maximize the surrogate loss under perturbation constraints:
\begin{equation}\small
	\max_{\bm{\phi} \in \mathcal{C}} \bigl\{\mathcal{L}_{\text{s}}(\bm{\phi}; \mathcal{S}_{\bm{\omega}}, \mathcal{D}_i) := \mathcal{L}_{\text{s}}(\mathcal{S}_{\bm{\omega}}(\boldsymbol{u}_i + \bm{\phi}), \boldsymbol{v}_i)\bigr\},
\end{equation}
where the perturbation $\bm{\phi}$ is constrained within an $\epsilon$-bounded $\ell_q$ norm ball $\mathcal{C} = \{\bm{\phi} \mid \|\bm{\phi}\|_q \leq \epsilon\}$, and  is typically generated via $K$-step projected gradient updates:
\begin{equation}\small
	\bm{\phi}_{k+1} \leftarrow \Pi_{\mathcal{C}}\left(\bm{\phi}_k + \alpha \cdot \text{sgn}(\nabla_{\bm{\phi}} \mathcal{L}_{\text{s}}(\bm{\phi}_k; \mathcal{S}_{\bm{\omega}}, \mathcal{D}_i))\right),
\end{equation}
for $k = 0, 1, \cdots, K-1$, where $\alpha$ is the step size, $\Pi_{\mathcal{C}}$ denotes projection onto $\mathcal{C}$, and $\text{sgn}(\cdot)$ is the sign operation. We denote IP by $\bm{\delta}$ (i.e., $\bm{\phi}_0 = \bm{\delta}$), which serves as the seed of the perturbation trajectory.  Such \emph{single-level} optimization neglects the interplay among different variables, limiting its generalization capability.

\subsection{Bilevel-Minimax Adversarial Transfer}

As shown in Fig.~\ref{fig:visual_3}, \texttt{BMAT} departs from the VTA paradigm by explicitly modeling a ternary coupling among these variables under a bilevel-minimax hierarchy. We next formalize this formulation and derive the bottom-up solver.

\subsubsection{Bilevel with Minimax Formulation}
We cast transfer attacks into a bilevel optimization problem whose inner subproblem adopts a minimax structure.

\textbf{Inner Minimax Modeling.} During attack iterations, perturbations that exploit universal gradient features of robust surrogates tend to exhibit enhanced transferability across victim models~\cite{pedraza2021relationship}. 
Motivated by this, \textcolor{black}{\texttt{BMAT}} first introduces an inner minimax objective, i.e., $f$ over $\boldsymbol{\phi}$ and $\boldsymbol{\omega}$ to capture their interaction:
\begin{equation} \small
	\min_{\boldsymbol{\phi}\in\mathcal{C}}\max_{\boldsymbol{\omega}\in\Omega}
	\bigl\{f(\boldsymbol{\phi}, \boldsymbol{\omega})
	:=-\mathcal{L}_{\textrm{s}}(\boldsymbol{\phi},\mathcal{S}_{\boldsymbol{\omega}};\mathcal{D}_{i})
	- \tau\,\mathcal{R}(\mathcal{S}_{\boldsymbol{\omega}};\mathcal{D}_{i})\bigr\},
	\label{eq:jma}
\end{equation} 
where $\mathcal{R}(\mathcal{S}_{\boldsymbol{\omega}};\mathcal{D}_{i})
:=\mathcal{L}_{\text{s}}(\mathcal{S}_{\bm{\omega}}(\boldsymbol{u}_i), \boldsymbol{v}_i)$ serves as a natural-accuracy regularizer with respect to $\boldsymbol{\omega}$, 
and $\tau>0$ is the  coefficient to balance robustness and natural accuracy.
Unlike VTA, which optimizes $\boldsymbol{\phi}$ against a fixed pretrained surrogate, this formulation jointly adapts both surrogate and perturbation to cultivate universal gradient features for cross-architecture transferability.

\textbf{Bilevel Modeling.} IP critically shapes the attack trajectory and the eventual transferability of
$\boldsymbol{\phi}$. VTA only evaluates transferability implicitly through the surrogate loss, which can cause AEs to overfit surrogate-specific artifacts. Instead of choosing $\boldsymbol{\delta}$ heuristically, \texttt{BMAT} optimizes IP using feedback from a pseudo-surrogate model $\mathcal{P}$. In practice, $\mathcal{P}$ is instantiated by reusing the available white-box surrogate (or its Bayesian version~\cite{li2023making}), without introducing any extra victim access, thus keeping the threat model identical to VTA. Let  $\mathcal{L}_{\text{p}}$ denote the loss w.r.t. $\mathcal{P}$. We then formulate the following bilevel–minimax problem:
\begin{equation}\small
	\begin{aligned}
		~~~~~\underset{\boldsymbol{\delta} \in \mathcal{C}}{\operatorname{min}}~ \bigl\{F(\boldsymbol{\delta},\boldsymbol{\phi}^{*}(\boldsymbol{\delta}))
		:=& -\mathcal{L}_{\text{p}}\bigl(\boldsymbol{\phi}^{*}(\boldsymbol{\delta}); \mathcal{P}, \mathcal{D}_{i} \bigr)\bigr\}, \\
		~~~\mathtt{where}~
		\boldsymbol{\phi}^{*}(\boldsymbol{\delta}) :=  \arg\min_{\boldsymbol{\phi} \in \mathcal{C}}&~ \max_{\boldsymbol{\omega} \in \Omega}~ f(\boldsymbol{\phi}, \boldsymbol{\omega}),\text{s.t.},~\boldsymbol{\phi}_0 = \boldsymbol{\delta}.
	\end{aligned}
\end{equation}
Here $\boldsymbol{\phi}^*(\boldsymbol{\delta})$ denotes the finite-step inner response induced by Eq.~\eqref{eq:jma} when the attack trajectory is \emph{seeded} at $\boldsymbol{\phi}_0 = \boldsymbol{\delta}$, rather than an exact global minimizer. Thus, the outer objective evaluates this local trajectory response on $\mathcal{P}$ and updates $\boldsymbol{\delta}$ accordingly. This view provides a principled mechanism for learning IP that consistently leads to stronger transfer attacks, with (i) the inner level performing robustness-aware adaptation of $(\boldsymbol{\phi},\boldsymbol{\omega})$ and (ii) the outer level optimizing $\boldsymbol{\delta}$ for trajectory seeding and improved transferability.

\begin{wrapfigure}{r}{0.55\textwidth}
	\vspace{-2.5em}
	\begin{minipage}{0.55\textwidth}
		\scriptsize
		\setlength{\intextsep}{0pt}
		\setlength{\columnsep}{2.5em}
		\setlength{\textfloatsep}{0pt}
		\setlength{\floatsep}{0pt}
		\SetAlCapFnt{\scriptsize}
		\SetAlCapNameFnt{\scriptsize}
		\begin{algorithm}[H]
			\DontPrintSemicolon
			\SetKwInOut{Input}{Input}
			\Input{Outer step $T$, inner step $\tilde{K}$, attack step $K$, surrogate $\S_{\Omegga}$, pseudo-surrogate $\mathcal{P}$
			}
			
			\textcolor{cvprblue}{\tcp{\textbf{I: Learning IP}}}
			\textbf{Initialize:} IP $\boldsymbol{\delta}^0$, pretrained weights $\Omegga_0$ \;
			
			\For{$t = 0$ \KwTo $T-1$}{
				$\boldsymbol{\phi}_0 \gets \boldsymbol{\delta}^t$ \textcolor{cvprblue}{\tcp*{\textbf{Trajectory Seeding}}}
				
				\textcolor{cvprblue}{\tcp{\textbf{Soft Weight Modulator}}}
				\For{$k = 0$ \KwTo $\tilde{K} - 1$}{
					Compute $\nabla_{\boldsymbol{\phi}}f_k$, $\nabla_{\boldsymbol{\omega}}f_k$ \;
					Update $\boldsymbol{\phi}_k$, $\boldsymbol{\omega}_k$ via Eq.~\eqref{eq:swm}
				}
				
				\textcolor{cvprblue}{\tcp{\textbf{Implicit Gradient Approximator}}}
				$\mathbf{h} \gets \texttt{\textbf{IGA}}\left(\nabla_{\boldsymbol{\phi}} F(\Phi_{\tilde{K}}),\ \nabla^2_{\boldsymbol{\phi}} f(\Phi_{\tilde{K}})\right)$ \;
				
				$\boldsymbol{\delta}^{t+1}\leftarrow\Pi_{\mathcal{C}} \left( \boldsymbol{\delta}^t - {\alpha} \cdot \mathtt{sgn}((\nabla^2_{\boldsymbol{\delta}\boldsymbol{\phi}} f(\Phi_{\tilde{K}}))^\top \mathbf{h}) \right)$ 
			}
			
			\textcolor{cvprblue}{\tcp{\textbf{II: Fast Transfer w/ Learned IP}}}
			$\boldsymbol{\phi}_0 \leftarrow \boldsymbol{\delta}^{T}$ \textcolor{cvprblue}{\tcp*{\textbf{Warm-Start}}}
			\For{$k = 0$ \KwTo $K-1$}{
				$\boldsymbol{\phi}_{k+1} \gets \Pi_{\mathcal{C}}\!\left(\boldsymbol{\phi}_k
				+ \alpha \cdot \mathtt{sgn}\big(\nabla_{\boldsymbol{\phi}}
				\mathcal{L}_{\text{s}}(\boldsymbol{\phi}_k; \mathcal{S}_{\Omegga}, \mathcal{D}_i)\big)\right)$ 
			}
			\KwRet Final attack $\boldsymbol{\phi}_K$
			
			\caption{\textbf{BMAT.}}
			\label{alg:bmat}
		\end{algorithm}
		
	\end{minipage}
\end{wrapfigure}

\subsubsection{Bottom-up Solver for \textbf{\texttt{BMAT}}}
We design  a bottom-up solver comprising SWM and IGA, aligned with the hierarchical problem structure.

\textbf{Soft Weight Modulator.} 
To efficiently solve Eq.~\eqref{eq:jma}, we introduce SWM to jointly update
$\boldsymbol{\phi}$ and $\boldsymbol{\omega}$ with a single backward pass.
Each outer iteration begins with $\bm{\phi}_0 = \bm{\delta}^t$, seeding the perturbation trajectory with the current initialization estimate.
SWM then performs $\tilde{K}$ inner steps to jointly optimize $(\boldsymbol{\phi}_k,\boldsymbol{\omega}_k)$ at low cost:

\begin{equation}\label{eq:swm}
	\small
	\setlength{\jot}{1pt} 
	\left\{
	\begin{aligned}
		\boldsymbol{\omega}_{k+1}\\[-2pt]
		\boldsymbol{\phi}_{k+1}
	\end{aligned}
	\right\}
	\leftarrow
	\left\{
	\begin{aligned}
		\boldsymbol{\omega}_{k} + \gamma \nabla_{\boldsymbol{\omega}} f_k\\[-2pt]
		\boldsymbol{\phi}_{k} - \beta \nabla_{\boldsymbol{\phi}} f_k
	\end{aligned}
	\right\}
\end{equation}
where $f_k := f(\boldsymbol{\phi}_k,\boldsymbol{\omega}_k)$.
Here the pretrained surrogate $\boldsymbol{\omega}_0$ serves as hard weights, and $\boldsymbol{\omega}_k$ are soft, adapted weights used within the inner loop. At the beginning of outer iteration (each batch), we restore $\boldsymbol{\omega}_0$ from the clean surrogate, so that SWM performs only local, attack-specific adaptation without accumulating cross-batch drift away from the original model. This design strengthens the perturbation using robustness-aware surrogate responses
without incurring extra backward passes compared to VTA.

\begin{wrapfigure}{r}{0.55\textwidth}		
	\vspace{-2.2em}
	\begin{minipage}{0.55\textwidth}
		\makeatletter
		\let\old@algocf@pre@ruled\@algocf@pre@ruled
		\makeatother
		\scriptsize
		\setlength{\intextsep}{0pt}
		\setlength{\columnsep}{1.9em}
		\setlength{\textfloatsep}{0pt}
		\setlength{\floatsep}{0pt}
				\SetAlCapFnt{\scriptsize}
				\SetAlCapNameFnt{\scriptsize}
		\begin{algorithm}[H]
			\SetKwInOut{Input}{Input}
			\DontPrintSemicolon
			\SetKwFunction{FCG}{\texttt{\textbf{IGA}}}
			\SetKwProg{Fn}{Function}{:}{}
			\Fn{\FCG{$\nabla_{\boldsymbol{\phi}} F$, $\nabla^2_{\boldsymbol{\phi}} f$}}{
				\Input{$\mathbf{h}_0 = 0$, tolerance $\zeta$, max iterations $N$}
				$r_0 \gets \nabla_{\boldsymbol{\phi}} F$, $p_0 \gets r_0$ \;
				\For{$\nu = 0,1,\dots,N$}{
					$\eta_\nu \gets \frac{r_\nu^\top r_\nu}{p_\nu^\top (\nabla^2_{\boldsymbol{\phi}} f \cdot p_\nu)}$ \;
					$\mathbf{h}_{\nu+1} \gets \mathbf{h}_\nu+\eta_\nu p_\nu,$
					$r_{\nu+1} \gets r_\nu-\eta_\nu (\nabla^2_{\boldsymbol{\phi}} f \cdot p_\nu)$ \;
					\If{$\|\nabla^2_{\boldsymbol{\phi}} f \cdot \mathbf{h}_{\nu+1} - \nabla_{\boldsymbol{\phi}} F\|_2 \leq \zeta$}{
						\KwRet $\mathbf{h}_{\nu+1}$
					}
					\textcolor{cvprblue}{\tcp{\textbf{Fletcher-Reeves Conjugation}}}
					$\lambda_\nu \gets \frac{r_{\nu+1}^\top r_{\nu+1}}{r_\nu^\top r_\nu},~$
					$p_{\nu+1} \gets r_{\nu+1} + \lambda_\nu p_\nu$
				}
				\KwRet $\mathbf{h}_N$
			}
			\caption{\textbf{IGA} for Computing $\mathbf{h}$.}	\label{alg:iga}
		\end{algorithm}
	\end{minipage}
\end{wrapfigure}

\textbf{Implicit Gradient Approximator.}  Central to the outer-level IP optimization is the
hypergradient calculation, i.e., $\nabla_{\boldsymbol{\delta}} F(\boldsymbol{\delta},\boldsymbol{\phi}^*(\boldsymbol{\delta}))$~\cite{liu2025augmenting,gao2024enhancing,liu2024learning,yue2024unveiling}. Whereas, unrolling the full inner minimax trajectory and backpropagating through
$\{\boldsymbol{\phi}_{k},\,\boldsymbol{\omega}_{k}\}_{k=1}^{\tilde K}$ incurs prohibitive computational and memory costs~\cite{liu2021investigating,gao2023learning,gao2026snoc}.
Instead, \texttt{BMAT} employs an implicit-gradient approximation to compute the hypergradient w.r.t. $\boldsymbol{\delta}$. By the implicit function theorem~\cite{liu2021investigating}, at the approximate inner optimum $\boldsymbol{\phi}^*(\boldsymbol{\delta})$, we have
\begin{equation}
	\small
	\nabla_{\boldsymbol{\delta}} F(\boldsymbol{\delta},\boldsymbol{\phi}^*(\boldsymbol{\delta}))
	=
	\bigl(\nabla_{\boldsymbol{\delta}\boldsymbol{\phi}}^2 f\bigr)^\top
	\bigl(\nabla_{\boldsymbol{\phi}\boldsymbol{\phi}}^2 f\bigr)^{-1}
	\nabla_{\boldsymbol{\phi}} F,
	\label{eq:hyper_exact}
\end{equation}
which can be equivalently rewritten as
\begin{equation}
	\small
	\nabla_{\boldsymbol{\delta}} F(\boldsymbol{\delta},\boldsymbol{\phi}^*(\boldsymbol{\delta}))
	\approx
	\bigl(\nabla_{\boldsymbol{\delta}\boldsymbol{\phi}}^2 f\bigr)^\top \mathbf{h},~
	\text{where } 
	\nabla_{\boldsymbol{\phi}\boldsymbol{\phi}}^2 f \,\mathbf{h} = \nabla_{\boldsymbol{\phi}} F.
	\label{eq:hyper_linear}
\end{equation}
Here $\mathbf{h}$ is the solution of a linear system defined by $\nabla_{\boldsymbol{\phi}\boldsymbol{\phi}}^2 f$ and $\nabla_{\boldsymbol{\phi}} F$. 
We denote the solver as $\texttt{\textbf{IGA}}(\nabla_{\boldsymbol{\phi}} F,\ \nabla^2_{\boldsymbol{\phi}} f)$, which returns $\mathbf{h}$ for approximating $\nabla_{\boldsymbol{\delta}}F(\Phi_{\tilde{K}}(\boldsymbol{\delta}))$, with details provided in Alg.~\ref{alg:iga}.
$\texttt{\textbf{IGA}}$ introduces a Fletcher-Reeves conjugate gradient~\cite{nocedal2006numerical},which avoids explicit Hessians and high-order backpropagation while respecting our bilevel-minimax coupling. $\mathbf{h}$ is then substituted into Eq.~\eqref{eq:hyper_linear} to update $\boldsymbol{\delta}$. Note that Eq.~\eqref{eq:hyper_exact} presents the idealized expression. 
In practice, we adopt a damped variant $(\nabla_{\phi\phi}^2 f + \rho I)^{-1}$ to ensure local invertibility and numerical stability. 

\textbf{Fast Transfer with IP.}
To attack unknown victim models, \texttt{BMAT} performs a standard iterative attack initialized at 
$\boldsymbol{\phi}_0 = \boldsymbol{\delta}^T$. The learned IP accelerates convergence and consistently yields stronger black-box transfer
than VTA. The complete procedure is summarized in Alg.~\ref{alg:bmat}.

\subsubsection{Algorithm Analysis}\label{sec:discussion}

The optimization dynamics of such bilevel-minimax formulations remain less understood in transfer-based attacks. Inspired by recent progress on bilevel optimization~\cite{hustochastic,liu2021investigating,yaoovercoming}, we provide a stability-oriented analysis of the regularized \texttt{BMAT} under the standard formulation: $\underset{\boldsymbol{\delta} \in \mathcal{C}}{\min} F\bigl(\boldsymbol{\delta}, \boldsymbol{\phi}^*(\boldsymbol{\delta}), 
\boldsymbol{\omega}^*(\boldsymbol{\delta})\bigr),$ where the inner response 
is defined by $
(\boldsymbol{\phi}^*(\boldsymbol{\delta}), \boldsymbol{\omega}^*(\boldsymbol{\delta}))=\arg \underset{\boldsymbol{\phi}\in\mathcal{C}}{\min} ~\underset{\boldsymbol{\omega}\in \Omega}{\max} 
f(\boldsymbol{\phi}, \boldsymbol{\omega}).
$

\begin{lemma}\label{lemma1}
		\small
	The following descent inequality holds:
	\small
	\begin{align*}
		F(\boldsymbol{\delta}_{t+1}, \boldsymbol{\phi}_{t+1,\tilde{K}}, \boldsymbol{\omega}_{t+1,\tilde{K}})
		\leq &~F(\boldsymbol{\delta}_t, \boldsymbol{\phi}_{t,\tilde{K}}, \boldsymbol{\omega}_{t,\tilde{K}}) -\frac{\alpha}{2}\bigl\|\nabla_{\boldsymbol{\delta}} 
		F(\boldsymbol{\delta}_t, \boldsymbol{\phi}_{t,\tilde{K}}, \boldsymbol{\omega}_{t,\tilde{K}})\bigr\|^2 \\
		&+ \frac{\alpha}{2}\,\epsilon_{\mathrm{IGA}}^{2}
		+ \frac{L_F \alpha^2}{2}\bigl(G_F + \epsilon_{\mathrm{IGA}}\bigr)^2.
	\end{align*}
\end{lemma}

\begin{lemma}\label{lemma2}
		\small
	The inner updates in \texttt{BMAT} (i.e., SWM) satisfy the following bounds:
	\small
	\begin{align*}
		\bigl\| \boldsymbol{\phi}_{t+1,\tilde{K}} 
		- \boldsymbol{\phi}^*(\boldsymbol{\delta}_{t+1})\bigr\|^2 
		&\leq 
		\bigl(1 + \beta^{2} L_{\boldsymbol{\phi}}^{2}\bigr)^{\tilde{K}} 
		\bigl\|\boldsymbol{\delta}_t 
		- \boldsymbol{\phi}^*(\boldsymbol{\delta}_{t+1})\bigr\|^2 
		+ \beta^{2} \tilde{K}\,\epsilon_{\boldsymbol{\phi}}^{2},\\[2pt]
		\bigl\|\boldsymbol{\omega}_{t+1,\tilde{K}}
		- \boldsymbol{\omega}^*(\boldsymbol{\delta}_{t+1})\bigr\|^2 
		&\leq 
		\bigl(1 + \gamma^{2} L_{\boldsymbol{\omega}}^{2}\bigr)^{\tilde{K}} 
		\bigl\|\boldsymbol{\omega}_0
		- \boldsymbol{\omega}^*(\boldsymbol{\delta}_{t+1})\bigr\|^2 
		+ \gamma^{2} \tilde{K}\,\epsilon_{\boldsymbol{\omega}}^{2}.
	\end{align*}
\end{lemma}
The following result characterizes the averaged stationarity behavior of the outer updates and provides insight into the stability of coupled optimization dynamics.
\begin{theorem}
		\small
	After running \texttt{BMAT} for $T$ iterations with step sizes 
	$\alpha = \beta = \gamma = \frac{c}{\sqrt{T}}$ 
	(where $c>0$ is a suitable constant), 
	there exist a constant $C > 0$ and error terms 
	$\epsilon_{\mathrm{IGA}},\epsilon^{(\tilde{K})}_{\boldsymbol{\phi}},
	\epsilon^{(\tilde{K})}_{\boldsymbol{\omega}}$ such that
	\small
	\begin{align*}
		\frac{1}{T+1} \sum_{t=0}^{T} 
		\bigl\|\nabla_{\boldsymbol{\delta}} 
		F(\boldsymbol{\delta}_t, \boldsymbol{\phi}_{t,\tilde{K}}, 
		\boldsymbol{\omega}_{t,\tilde{K}})\bigr\|^2 
		\leq \frac{C}{\sqrt{T}} 
		+ \epsilon_{\mathrm{IGA}}^{2} 
		+ \bigl(\epsilon^{(\tilde{K})}_{\boldsymbol{\phi}}\bigr)^{2} 
		+ \bigl(\epsilon^{(\tilde{K})}_{\boldsymbol{\omega}}\bigr)^{2}.
	\end{align*}
\end{theorem}

\section{Experiments} 
\textbf{Datasets and Models.} Experiments are conducted on ImageNet (classification) and Cityscapes and ADE20K (segmentation). For classification, ResNet-50 serves as the surrogate model. The 10 victims include 4 CNNs ([1]-[4]: IncRes-v2~\cite{szegedy2017inception}, MobileNet~\cite{sandler2018mobilenetv2}, PNASNet~\cite{liu2018progressive}, and SENet~\cite{hu2018squeeze}), 3 robust ensembles ([5]-[7]: Inc-v3$_\text{ens3}$, Inc-v3$_\text{ens4}$, IncRes-v2$_\text{ens}$)~\cite{tramer2017ensemble}, and 3 Transformers ([8]-[10]: ViT~\cite{dosovitskiy2020image},Visformer~\cite{chen2021visformer}, and Swin~\cite{liu2021swin}). Inception-v3~\cite{szegedy2016rethinking} is used as the ensemble model for MBA and pseudo-surrogate for \texttt{BMAT}. For segmentation, we use MMSegmentation~\cite{mmseg2020}. Each dataset includes 10 victims: 8 CNN-based models (FCN~\cite{shelhamer2017fully},  UPerNet~\cite{xiao2018unified}, PSANet~\cite{zhao2018psanet}, ISANet~\cite{huang2019isa}, DLV3-R50/R101~\cite{chen2017rethinking}, and PSP-R50/R101~\cite{wightman2021resnet}) and 2 Transformer-based models (Segformer~\cite{xie2021segformer} and Setr~\cite{zheng2020rethinking}). GCNet~\cite{cao2019gcnet} is used analogously for EBAD and \texttt{BMAT} in segmentation.

\begin{table}[!t]
	\centering
	\caption{ASR results of combining \texttt{BMAT} with 9 mainstream attack methods. Surrogates $[1]$-$[7]$ and$[8]$-$[10]$ denote CNN- and transformer-based victims. Performance gains in $\mathbf{(0,5]}$ and $>\mathbf{5}$ are marked with \colorbox{myblue}{light blue} and \colorbox{mydblue}{deep blue}, respectively.}
	\label{tab:asr_combined_results}
	\scalebox{0.6}{
		\begin{threeparttable}
			\renewcommand\arraystretch{1.4}
			\setlength{\tabcolsep}{0.45mm}{
				\begin{tabular}{ccccccccccccccccccccccccc}
					\toprule[1.2pt]
					\multicolumn{24}{c}{\normalsize Image Classification, ResNet-50 backbone, ImageNet dataset, ASR $\uparrow$} \\
					\cmidrule{1-13}\cmidrule{13-24}
					\multicolumn{2}{c}{\multirow{3}{*}{\shortstack{Basic \\ Attacker}}}&\multicolumn{4}{c}{CNN}&\multicolumn{3}{c}{CNN Ensemble}&\multicolumn{3}{c}{Transformer }& \multicolumn{2}{c}{\multirow{3}{*}{\shortstack{Basic \\ Attacker}}}&\multicolumn{4}{c}{CNN}&\multicolumn{3}{c}{CNN Ensemble}&\multicolumn{3}{c}{Transformer }\\
					\cmidrule(lr){3-6}\cmidrule(lr){7-9}\cmidrule(lr){10-12} \cmidrule(lr){15-18}\cmidrule(lr){19-21}\cmidrule(lr){22-24} 
					&&  \scriptsize $[1]$   &\scriptsize $[2]$ &   \scriptsize$[3]$ & \scriptsize$[4]$&\scriptsize $[5]$& \scriptsize$[6]$ & \scriptsize$[7]$ &\scriptsize $[8]$&\scriptsize $[9]$&\scriptsize $[10]$  &&&\scriptsize $[1]$   &\scriptsize $[2]$ &\scriptsize   $[3]$ &\scriptsize $[4]$&\scriptsize $[5]$&\scriptsize $[6]$ &\scriptsize $[7]$ &\scriptsize $[8]$& \scriptsize $[9]$& \scriptsize$[10]$  \\ \hline			
					\multirow{8}{*}{PGD} 
					& N/A & 13.64 & 36.58 & 13.28 & 16.84 & 10.34 & 9.38 & 5.58 & 4.22 & 11.04 & 12.8& \multirow{8}{*}{~~~SI} 
					& N/A & 17.08 & 46.26 & 17.76 & 22.84 & 13.28 & 11.42 & 6.84 & 5.56 & 15.26 & 16.32 \\ 
					& \textbf{Ours} & \cellcolor{mydblue}20.12 & \cellcolor{mydblue}47.52 & \cellcolor{mydblue}20.40 & \cellcolor{mydblue}24.06 & \cellcolor{myblue}14.04 & \cellcolor{myblue}13.08 & \cellcolor{myblue}7.62 & \cellcolor{myblue}5.76 & \cellcolor{myblue}15.46 & \cellcolor{myblue}16.28 && \textbf{Ours}& \cellcolor{mydblue}23.78 & \cellcolor{mydblue}55.18 & \cellcolor{mydblue}23.92 & \cellcolor{mydblue}30.28 & \cellcolor{myblue}16.60 & \cellcolor{myblue}14.02 & \cellcolor{myblue}8.38 & \cellcolor{myblue}6.50 & \cellcolor{myblue}19.10 & \cellcolor{myblue}19.20 \\ 
					& SGM & 20.06 & 55.74 & 21.92 & 29.94 & 13.86 & 12.68 & 7.62 & 8.06 & 21.36 & 24.18 && SGM& 27.18 & 66.34 & 30.34 & 38.68 & 18.16 & 15.38 & 9.92 & 9.92 & 28.78 & 30.50 \\ 
					& \textbf{Ours} & \cellcolor{mydblue}27.86 & \cellcolor{mydblue}63.36 & \cellcolor{mydblue}29.50 & \cellcolor{mydblue}37.34 & \cellcolor{myblue}18.32 & \cellcolor{myblue}16.18 & \cellcolor{myblue}10.06 & \cellcolor{myblue}9.62 & \cellcolor{myblue}26.00 & \cellcolor{myblue}27.44 && \textbf{Ours}& \cellcolor{mydblue}34.02 & \cellcolor{mydblue}72.86 & \cellcolor{mydblue}36.52 & \cellcolor{mydblue}44.20 & \cellcolor{myblue}21.24 & \cellcolor{myblue}18.20 & \cellcolor{myblue}11.48 & \cellcolor{myblue}11.04 & \cellcolor{myblue}32.26 & \cellcolor{myblue}32.72 \\ 
					& Ghost & 13.92 & 41.64 & 14.12 & 19.16 & 11.04 & 10.54 & 6.18 & 4.60 & 12.56 & 13.86 && Ghost & 18.82 & 50.94 & 19.44 & 25.50 & 13.86 & 12.48 & 7.34 & 5.18 & 16.62 & 16.54 \\ 
					& \textbf{Ours} & \cellcolor{mydblue}21.18 & \cellcolor{mydblue}51.94 & \cellcolor{mydblue}21.48 & \cellcolor{myblue}26.74 & \cellcolor{myblue}14.86 & \cellcolor{myblue}13.36 & \cellcolor{myblue}8.04 & \cellcolor{myblue}5.82 & \cellcolor{myblue}15.94 & \cellcolor{myblue}17.52 && \textbf{Ours}& \cellcolor{mydblue}24.78 & \cellcolor{mydblue}59.38 & \cellcolor{mydblue}25.20 & \cellcolor{mydblue}31.26 & \cellcolor{myblue}17.12 & \cellcolor{myblue}14.04 & \cellcolor{myblue}8.04 & \cellcolor{myblue}6.06 & \cellcolor{myblue}19.34 & \cellcolor{myblue}18.78 \\ 
					& MBA & 15.34 & 57.22 & 14.04 & 20.40 & 11.10 & 10.26 & 6.26 & 3.78 & 12.22 & 13.28 && MBA& 20.64 & 64.18 & 19.18 & 25.48 & 15.88 & 12.90 & 8.74 & 4.14 & 14.62 & 14.66 \\ 
					& \textbf{Ours} & \cellcolor{mydblue}20.98 & \cellcolor{mydblue}64.26 & \cellcolor{mydblue}19.90 & \cellcolor{mydblue}25.96 & \cellcolor{myblue}15.52 & \cellcolor{myblue}13.30 & \cellcolor{myblue}8.56 & \cellcolor{myblue}4.62 & \cellcolor{myblue}15.20 & \cellcolor{myblue}15.52 && \textbf{Ours}& \cellcolor{mydblue}26.82 & \cellcolor{mydblue}71.08 & \cellcolor{mydblue}24.68 & \cellcolor{mydblue}31.44 & \cellcolor{myblue}19.26 & \cellcolor{myblue}15.88 & \cellcolor{myblue}10.44 & \cellcolor{myblue}5.10 & \cellcolor{myblue}17.44 & \cellcolor{myblue}17.40 \\ 
					\hline
					\multirow{8}{*}{MI} 
					& N/A & 22.02 & 49.70 & 22.92 & 30.24 & 15.88 & 14.46 & 8.82 & 7.54 & 18.40 & 18.92 & \multirow{8}{*}{~~~TI} &N/A& 14.36 & 38.32 & 15.94 & 19.82 & 10.98 & 10.48 & 6.4 & 5.34 & 11.22 & 11.86 \\ 
					& \textbf{Ours} & \cellcolor{mydblue}34.78 & \cellcolor{mydblue}61.50 & \cellcolor{mydblue}34.84 & \cellcolor{mydblue}38.66 & \cellcolor{mydblue}25.92 & \cellcolor{mydblue}22.62 & \cellcolor{mydblue}14.72 & \cellcolor{myblue}10.14 & \cellcolor{mydblue}23.98 & \cellcolor{mydblue}24.34 && \textbf{Ours}& \cellcolor{mydblue}19.54 & \cellcolor{mydblue}47.96 & \cellcolor{mydblue}21.90 & \cellcolor{mydblue}25.24 & \cellcolor{myblue}14.90 & \cellcolor{myblue}13.00 & \cellcolor{myblue}8.26 & \cellcolor{myblue}5.86 & \cellcolor{myblue}13.54 & \cellcolor{myblue}13.98 \\ 
					& SGM & 28.76 & 66.00 & 31.64 & 41.40 & 19.62 & 17.44 & 11.30 & 11.60 & 28.38 &28.90&& SGM& 21.96 & 57.62 & 26.92 & 33.92 & 15.68 & 14.04 & 8.94 & 9.66 & 22.62 & 24.24 \\ 
					& \textbf{Ours} & \cellcolor{mydblue}38.74 & \cellcolor{mydblue}72.96 & \cellcolor{mydblue}40.36 & \cellcolor{mydblue}47.30 & \cellcolor{mydblue}27.44 & \cellcolor{mydblue}23.96 & \cellcolor{myblue}16.02 & \cellcolor{myblue}13.14 & \cellcolor{myblue}32.22 & \cellcolor{myblue}32.64 && \textbf{Ours}& \cellcolor{mydblue}28.24 & \cellcolor{mydblue}64.48 & \cellcolor{mydblue}32.56 & \cellcolor{mydblue}39.46 & \cellcolor{myblue}19.00 & \cellcolor{myblue}16.76 & \cellcolor{myblue}10.62 & \cellcolor{myblue}10.26 & \cellcolor{myblue}25.32 & \cellcolor{myblue}25.96 \\ 
					& Ghost & 23.60 & 56.20 & 24.70 & 34.10 & 17.24 & 15.24 & 9.62 & 7.44 & 20.30 & 20.04&& Ghost &  15.66 & 43.02 & 17.02 & 22.74 & 12.56 & 10.96 & 6.86 & 4.96 & 12.68 & 12.54 \\ 
					& \textbf{Ours} & \cellcolor{mydblue}38.72 & \cellcolor{mydblue}68.58 & \cellcolor{mydblue}38.66 & \cellcolor{mydblue}43.82 & \cellcolor{mydblue}28.40 & \cellcolor{mydblue}24.36 & \cellcolor{mydblue}16.14 & \cellcolor{myblue}10.12 & \cellcolor{mydblue}26.34 & \cellcolor{mydblue}26.44&& \textbf{Ours} & \cellcolor{myblue}19.78 & \cellcolor{mydblue}52.18 & \cellcolor{mydblue}22.16 & \cellcolor{myblue}27.70 & \cellcolor{myblue}15.02 & \cellcolor{myblue}12.74 & \cellcolor{myblue}7.78 & \cellcolor{myblue}5.64 & \cellcolor{myblue}14.18 & \cellcolor{myblue}13.96 \\ 
					& MBA & 28.34 & 76.52 & 28.98 & 36.48 & 21.10 & 17.62 & 11.70 & 6.58 & 20.82 &20.62&& MBA& 18.48 & 60.86 & 19.36 & 24.06 & 15.62 & 13.28 & 8.68 & 4.82 & 12.56 & 12.84 \\ 
					& \textbf{Ours} & \cellcolor{mydblue}39.54 & \cellcolor{myblue}80.66 & \cellcolor{mydblue}37.90 & \cellcolor{mydblue}43.02 & \cellcolor{mydblue}30.74 & \cellcolor{mydblue}26.84 & \cellcolor{mydblue}17.36 & \cellcolor{myblue}9.18 & \cellcolor{myblue}25.52 & \cellcolor{myblue}24.66&& \textbf{Ours} & \cellcolor{mydblue}24.92 & \cellcolor{mydblue}67.28 & \cellcolor{mydblue}26.94 & \cellcolor{mydblue}30.46 & \cellcolor{myblue}19.84 & \cellcolor{myblue}16.66 & \cellcolor{myblue}11.24 & \cellcolor{myblue}5.68 & \cellcolor{myblue}15.76 & \cellcolor{myblue}15.00 \\ 
					\hline
					\multirow{8}{*}{VMI} 
					& N/A & 31.04 & 60.16 & 33.62 & 39.94 & 23.12 & 20.88 & 14.02 & 10.74 & 25.62 & 26.14 & \multirow{8}{*}{~~~DI} &N/A& 31.0 & 62.8 & 36.66 & 38.66 & 23.0 & 20.58 & 13.5 & 8.24 & 21.54 & 20.78 \\ 
					& \textbf{Ours} & \cellcolor{mydblue}44.06 & \cellcolor{mydblue}69.46 & \cellcolor{mydblue}44.46 & \cellcolor{mydblue}47.02 & \cellcolor{mydblue}33.28 & \cellcolor{mydblue}29.42 & \cellcolor{mydblue}20.14 & \cellcolor{myblue}13.08 & \cellcolor{mydblue}31.42 & \cellcolor{myblue}31.10 && \textbf{Ours}& \cellcolor{mydblue}38.54 & \cellcolor{mydblue}70.12 & \cellcolor{mydblue}41.84 & \cellcolor{mydblue}45.06 & \cellcolor{myblue}26.88 & \cellcolor{myblue}22.82 & \cellcolor{myblue}14.8 & \cellcolor{myblue}9.06 & \cellcolor{myblue}24.84 & \cellcolor{myblue}23.3 \\ 
					& SGM & 38.36 & 74.46 & 42.34 & 50.34 & 27.70 & 24.36 & 16.16 & 15.32 & 36.38  & 37.12& & SGM& 42.66 & 79.8 & 48.7 & 52.96 & 29.9 & 25.5 & 17.86 & 14.22 & 36.42 & 36.14 \\ 
					& \textbf{Ours} & \cellcolor{mydblue}48.70 & \cellcolor{myblue}78.80 & \cellcolor{mydblue}49.88 & \cellcolor{myblue}54.74 & \cellcolor{mydblue}36.16 & \cellcolor{mydblue}31.08 & \cellcolor{mydblue}21.42 & \cellcolor{myblue}16.60 & \cellcolor{myblue}39.56 & \cellcolor{myblue}40.22 && \textbf{Ours}& \cellcolor{mydblue}49.74 & \cellcolor{myblue}84.48 & \cellcolor{mydblue}55.4 & \cellcolor{mydblue}58.6 & \cellcolor{myblue}34.74 & \cellcolor{myblue}28.5 & \cellcolor{myblue}19.98 & \cellcolor{myblue}15.72 & \cellcolor{myblue}38.96 & \cellcolor{myblue}37.96 \\ 
					& Ghost & 32.50 & 65.82 & 35.10 & 42.54 & 24.86 & 21.42 & 14.56 & 10.38 & 26.74 & 27.02 && Ghost& 31.5 & 66.32 & 34.76 & 39.5 & 23.68 & 20.06 & 12.68 & 7.76 & 20.9 & 20.62 \\ 
					& \textbf{Ours} & \cellcolor{mydblue}47.16 & \cellcolor{mydblue}73.72 & \cellcolor{mydblue}47.34 & \cellcolor{mydblue}50.44 & \cellcolor{mydblue}35.86 & \cellcolor{mydblue}30.44 & \cellcolor{mydblue}20.26 & \cellcolor{myblue}12.86 & \cellcolor{myblue}33.48 & \cellcolor{mydblue}33.00 && \textbf{Ours}& \cellcolor{mydblue}37.18 & \cellcolor{mydblue}71.74 & \cellcolor{mydblue}40.36 & \cellcolor{mydblue}44.68 & \cellcolor{myblue}25.36 & \cellcolor{myblue}21.44 & \cellcolor{myblue}13.6 & \cellcolor{myblue}7.9 & \cellcolor{myblue}22.96 & \cellcolor{myblue}22.74 \\ 
					& MBA & 33.50 & 81.56 & 34.80 & 40.70 & 25.96 & 22.44 & 14.74 & 8.22 & 24.72 & 24.98 && MBA& 23.4 & 66.86 & 20.7 & 28.72 & 19.4 & 16.62 & 10.38 & 4.28 & 12.42 & 11.76 \\ 
					& \textbf{Ours} & \cellcolor{mydblue}42.50 & \cellcolor{myblue}82.26 & \cellcolor{mydblue}40.84 & \cellcolor{myblue}44.94 & \cellcolor{mydblue}33.52 & \cellcolor{mydblue}29.12 & \cellcolor{myblue}18.70 & \cellcolor{myblue}10.00 & \cellcolor{myblue}27.10 & \cellcolor{myblue}26.56 && \textbf{Ours}& \cellcolor{mydblue}30.8 & \cellcolor{mydblue}71.88 & \cellcolor{mydblue}26.74 & \cellcolor{mydblue}34.2 & \cellcolor{myblue}24.18 & \cellcolor{myblue}20.66 & \cellcolor{myblue}13.36 & \cellcolor{myblue}5.1 & \cellcolor{myblue}15.4 & \cellcolor{myblue}13.82 \\ 
					\bottomrule[1.2pt]
				\end{tabular}
			}
		\end{threeparttable}
	}
\end{table}

\noindent \textbf{Baselines and Evaluation Metrics.} We report ASR ($\uparrow$) for image classification and mIoU ($\downarrow$) for semantic segmentation tasks. We consider 12 mainstream attackers, including standard PGD~\cite{kurakin2018adversarial}, model-based SGM~\cite{wu2020boosting} and Ghost~\cite{li2020learning}, input-transformation-based SI~\cite{lin2019nesterov}, DI~\cite{xie2019improving} and TI~\cite{dong2019evading}, momentum-based MI~\cite{dong2018boosting}, VMI~\cite{wang2021enhancing}, GMI~\cite{wang2024boosting}, and RAP~\cite{qin2022boosting}, ensemble-based MBA~\cite{li2023making}, initialization learning based BETAK~\cite{liu2024advancing}, and surrogate-adaptation-based DRA~\cite{zhu2022toward} and FAUG~\cite{wang2025improving}. 
For segmentation, we further include momentum-based NI~\cite{lin2019nesterov}, SegPGD~\cite{gu2022segpgd}, ensemble-based EBAD~\cite{cai2023ensemble}, and CosPGD~\cite{agnihotri2024cospgd}. 
These baselines span diverse categories to ensure comprehensive evaluation of \texttt{BMAT}.

\subsection{Experimental Results}

\textbf{Image Classification.} Tab.~\ref{tab:asr_combined_results} presents the ASR results by enhancing $9$ base attackers with \texttt{BMAT}, resulting in $24$ variants to evaluate its flexibility. As shown, whether built on CNN or Transformers, or ensemble-based robust models, the AEs generated with \texttt{BMAT} exhibit consistently stronger generalizability across $10$ victims, yielding an average ASR gain of {23.28}\% across 24 combinations. 

\begin{table}[!t]
	\centering
	\caption{Analysis of different initialization types and  surrogate adaptation techniques.} 
	\label{tab:gmi}
	\scalebox{0.7}{
		\begin{threeparttable}
			\renewcommand\arraystretch{1.4}
			\setlength{\tabcolsep}{0.1mm}{
				\begin{tabular}{ccccccccccccc}
					\toprule[1.2pt]
					\multicolumn{12}{c}{ ImageNet dataset, ResNet-50 backbone, ASR $\uparrow$} \\
					\cmidrule{1-12}
					\multicolumn{2}{c}{\multirow{3}{*}{\shortstack{Basic \\ Attacker}}}&\multicolumn{4}{c}{CNN}&\multicolumn{3}{c}{CNN Ensemble}&\multicolumn{3}{c}{Transformer }\\
					\cmidrule(lr){3-6}\cmidrule(lr){7-9}\cmidrule(lr){10-12}  
					&  & \scriptsize $[1]$ IncRes-V2   &\scriptsize  $[2]$ MobileNet&\scriptsize $[3]$ PNASNet &\scriptsize $[4]$ SENet&\scriptsize $[5]$ Inc-v$3_{ens3}$ &\scriptsize $[6]$ Inc-v$3_{ens4}$ &\scriptsize $[7]$ IncRes-v$2_{ens }$ &\scriptsize $[8]$ ViT&\scriptsize $[9]$ Visformer&\scriptsize $[10]$ Swin  \\ \hline                                                   
					\multirow{3}{*}{MI}& N/A
					&  22.02 &  49.70 &  22.92 &  30.24 &  15.88 &  14.46 &  8.82 &  7.54 &  18.40 &  18.92 \\ 
					&\multicolumn{1}{c}{  GMI}&  25.70 &  57.84 &  26.02 &  35.92 &  17.16 &  15.56 &  9.76 &  7.84 &  21.02 &  21.86  \\ 
					&\multicolumn{1}{c}{   \textbf{Ours}}&  \cellcolor{mydblue} 35.78&  \cellcolor{mydblue} 64.38 &  \cellcolor{mydblue} 35.40 &  \cellcolor{mydblue} 42.62 &  \cellcolor{mydblue} 24.84 &  \cellcolor{mydblue} 21.58 &  \cellcolor{myblue} 14.26 &  \cellcolor{myblue} 10.38 &  \cellcolor{myblue} 25.32&  \cellcolor{myblue} 25.92 \\ 
					\hline
					\multirow{3}{*}{VMI}& N/A
					&  31.04&  60.16 &  33.62 &  39.94 &  23.12 &  20.88 &  14.02 &  10.74 &  25.62 &  26.14\\ 
					&\multicolumn{1}{c}{  GMI}&  32.98 &  64.92 &  35.10 &  43.22 &  23.74 &  20.92 &  13.66 &  10.46 &  26.80 &  26.64  \\ 
					&\multicolumn{1}{c}{  \textbf{Ours}} &  \cellcolor{mydblue} 45.36&  \cellcolor{mydblue} 72.78 &  \cellcolor{mydblue} 45.84 &  \cellcolor{mydblue} 50.74&  \cellcolor{mydblue} 33.62 &  \cellcolor{mydblue} 28.86 &  \cellcolor{mydblue} 19.86 &  \cellcolor{myblue} 13.24&  \cellcolor{mydblue} 32.98 &  \cellcolor{mydblue} 32.30  \\ 
					\hline
					\multirow{4}{*}{PGD}&\multicolumn{1}{c}{ DRA}&  20.34 &  56.68 &  17.66 &  15.54 &  25.04 &  26.16 &  17.60 &  5.32&  9.86 &  10.48  \\ 
					&\multicolumn{1}{c}{ \bf Ours} &  \cellcolor{myblue} 21.84&  \cellcolor{myblue} 58.38 &  \cellcolor{myblue} 19.06 &  \cellcolor{myblue} 16.62&  \cellcolor{myblue} 26.14 &  \cellcolor{myblue} 27.5 &  \cellcolor{myblue} 18.80 &  \cellcolor{myblue} 5.62&  \cellcolor{myblue} 10.30&  \cellcolor{myblue} 10.70 \\ 
					&\multicolumn{1}{c}{ FAUG}&  21.70 &  32.34 &  22.96 &  21.02 &  15.42 &  14.20 &  10.60 &  8.24&  18.38 &  19.62  \\ 
					&\multicolumn{1}{c}{  \bf Ours} &  \cellcolor{myblue} 25.48&  \cellcolor{mydblue} 37.34 &  \cellcolor{myblue} 27.52 &  \cellcolor{myblue} 25.06&  \cellcolor{myblue} 19.60 &  \cellcolor{myblue} 17.74 &  \cellcolor{myblue} 12.42 &  \cellcolor{myblue} 9.06&  \cellcolor{myblue} 21.12 &  \cellcolor{myblue} 22.02  \\ 
					\bottomrule[1.2pt]
				\end{tabular}
			}
		\end{threeparttable}
	}
\end{table}

\begin{table}[!t]
	\centering
	\caption{Comparative results under normalized budgets, i.e., Backward Passes (BP). The best results are denoted with \textbf{boldface}.}
	\label{tab:betak}
	\scalebox{0.7}{
		\begin{threeparttable}
			\renewcommand\arraystretch{1.4}
			\setlength{\tabcolsep}{3.2mm}
			\begin{tabular}{lcccccc}
				\toprule[1.2pt]
				\multirow{1}{*}{Method} & \multirow{1}{*}{CNN} & \multirow{1}{*}{CNN Ensemble} & \multirow{1}{*}{Transformer} & Avg. ASR & Memory (GB)& Runtime  (Sec) \\
				\hline
				PGD (BP=10) & 10.93 & 5.73 & 7.16 & 8.24 & \textbf{3.21} & \textbf{2.68} \\
				PGD (BP=40) & 11.05 & 5.25 & 6.70 & 8.00 & 3.22 & 3.15 \\
				\hline
				RAP (BP=40) & 7.31 & 4.67 & 3.71 & 5.44 & 3.75 & 3.07\\
				RAP (BP=400) & 13.74 & 6.46 & 7.47 & 9.68 & 5.46 & 6.91 \\
				\hline
				BETAK (BP=40) & 17.16 & 8.07 & 9.25 & 12.06 & 22.69 & 6.37 \\
				\hline
				\textbf{Ours (BP=40)} & \textbf{22.52} & \textbf{8.75} & \textbf{11.34} & \textbf{15.03} & 7.89 & 4.64 \\
				\bottomrule[1.2pt]
			\end{tabular}
		\end{threeparttable}
	}
\end{table}

\textbf{Comparison with Initialization and Adaptation Techniques.} Vanilla momentum-based attackers use zero initialization, while GMI adds a global momentum warm start on a fixed surrogate.  As shown in Tab.~\ref{tab:gmi}, across all victims, \texttt{BMAT} consistently yields higher ASR than both the vanilla and GMI variants. We also compare with surrogate-adaptation attacks, i.e., DRA and FAUG. \texttt{BMAT} also achieves uniformly better transfer results, demonstrating the effectiveness of the bilevel-minimax coordination.

\textbf{Comparison with Bilevel and Minimax-based Attacks.} We further compare \texttt{BMAT} with RAP and BETAK, which are more closely related in formulation. RAP enhances transferability via repeated explicit maximization within a single-level minimax framework. 
In contrast, \texttt{BMAT} models the interaction among initialization, perturbation, and surrogate adaptation in a unified bilevel-minimax formulation. 
As shown in Tab.~\ref{tab:betak}, under normalized computational budgets (BP=40), \texttt{BMAT} consistently achieves higher ASR than RAP, and remains superior even when RAP increases its budget by $10\times$. 
Compared with BETAK, which relies on ensemble-guided initialization updates, \texttt{BMAT} achieves stronger transferability with substantially lower memory overhead, indicating the benefit of unified coordination without ensemble dependence.


\textbf{Semantic Segmentation.} In Tab.~\ref{tab:atk_1}, we present mIoU comparisons based on $3$ surrogate structures. Momentum-based methods notably improve AE transferability. In comparison, although EBAD incorporates gradients from additional surrogate models, it struggles to produce perturbations with stronger generalizability. In contrast, \texttt{BMAT} consistently enhances transferability across diverse models. Notably, with Segformer as the surrogate, \texttt{BMAT} surpasses other attackers with nearly $2\times $ higher transferability.

\textbf{Discussion on Transferability.}
The segmentation results further show that stronger white-box optimization does not necessarily yield stronger black-box transfer. Momentum-based attacks already perform well when the surrogate and victim share similar CNN structures, but their gains are less stable on transformer victims. \texttt{BMAT} improves this harder cross-architecture regime by adapting the attack trajectory instead of merely strengthening perturbation updates on a fixed surrogate.

\begin{table}[!t]		
	\centering
	\caption{Comparison with 6 mainstream attacks on segmentation tasks. $\dag$, $\ddag$, and $*$ denote the white-box surrogate models used for generating AEs. The best and second-best results are designated with $\textbf{boldface}$ and \underline{underline}, respectively. } 
	\label{tab:atk_1}
	\scalebox{0.7}{
		\begin{threeparttable} 
			\renewcommand\arraystretch{1.3}
			\setlength{\tabcolsep}{0.5mm}{
				\begin{tabular}{cccccccccccc}
					\toprule[1.2pt] 
					\multicolumn{12}{c}{ \normalsize Semantic Segmentation, Cityscapes dataset, mIoU $\downarrow$}\\
					\cmidrule{1-12}
					\multirow{2}{*}{\shortstack{Surrogate \\ Model}}&	\multirow{2}{*}{\shortstack{Basic \\ Attacker}}&\multicolumn{8}{c}{CNN }&\multicolumn{2}{c}{Transformer}\\
					\cmidrule(lr){3-10}\cmidrule(lr){11-12}  
					&  & FCN$^{\dag}$& UPerNet & PSANet & ISANet  &DLV3-R50$^{\ddag}$ &DLV3-R101&PSP-R50&PSP-R101 &Segformer$^{*}$& Setr   \\
					\hline
					\multicolumn{2}{c}{ Clean Data}&72.25&77.10&77.63&78.49&79.09&80.20&77.85&78.34&76.54&78.10 \\
					\hline \multirow{6}{*}{ FCN$^{\dag}$ } & PGD &1.97$^{\dag}$&3.74&3.43&3.31&3.52&8.09&3.61&6.83&33.96&42.09\\
					& SegPGD&2.02$^{\dag}$&3.60&\underline{3.33}&3.05&\underline{3.09}&10.04&\underline{3.20}&7.98&36.65&44.73 \\
					& CosPGD & $2.63^{\dag}$& 5.74& 5.10& 4.50 & 4.96 & 13.42& 5.37&10.40 & 35.30 & 42.79  \\ 
					& NI &\underline{1.89}$^{\dag}$&3.92&3.76&\underline{2.77}&3.58&7.23&3.28&5.79&29.40&40.43\\
					& MI &2.40$^{\dag}$&\underline{3.57}&3.38&3.17&3.55&\underline{6.52}&3.51&\underline{5.39}&\underline{27.81}&\underline{38.75}\\
					
					& EBAD &2.00$^{\dag}$&3.83&3.55&3.34&3.54&8.40&3.81&7.05&33.91&42.10\\
					&  \textbf{Ours} & \cellcolor{myblue} $\mathbf{1.75}^{\dag}$ &\cellcolor{myblue}  $\mathbf{2.74}$& \cellcolor{myblue} $\mathbf{2.60}$& \cellcolor{myblue}  $\mathbf{2.42}$ & \cellcolor{myblue}  $\mathbf{2.81}$ &  \cellcolor{myblue} $\mathbf{5.30}$& \cellcolor{myblue}  $\mathbf{2.62}$& \cellcolor{myblue}  $\mathbf{4.44}$ & \cellcolor{myblue}  $\mathbf{26.58}$& \cellcolor{myblue} $\mathbf{38.35}$\\ 
					\hline
					\multirow{6}{*}{  \shortstack{DLV3-R50$^{\ddag}$} } & PGD &7.55&4.85&2.74&4.78&$\underline{0.81}$$^{\ddag}$&15.02&\underline{3.04}&11.43&41.04&48.37\\
					& SegPGD &10.43&6.20&3.74&4.43&1.08$^{\ddag}$&18.02&4.40&14.45&43.57&49.74 \\
					& CosPGD &10.57&6.34&3.94&5.23&\cellcolor{myblue} $\mathbf{0.64}^{\ddag}$&20.31&4.99&14.75&41.78&48.73\\
					& NI &5.88&4.80&5.41&4.34&2.11$^{\ddag}$&9.88&4.14&6.77&33.82& 45.19\\
					& MI &\underline{5.57}&\underline{4.29}&2.95&\underline{4.05}&0.99$^{\ddag}$&\underline{9.55}&3.77&\underline{6.52}&\underline{32.99}&\underline{43.38}\\
					
					& EBAD &7.42&4.90&\underline{2.62}&4.83&0.77$^{\ddag}$&15.45&3.20&11.50&41.16& 48.45\\
					&\textbf{Ours} & \cellcolor{myblue}  $\mathbf{2.60}$ &  \cellcolor{myblue} $\mathbf{1.96}$&  \cellcolor{myblue} $\mathbf{2.16}$& \cellcolor{myblue}  $\mathbf{2.31}$ & $1.34$$^{\ddag}$ &\cellcolor{mydblue}   $\mathbf{4.54}$& \cellcolor{myblue}  $\mathbf{1.86}$&  \cellcolor{myblue} $\mathbf{3.57}$ & \cellcolor{myblue}  $\mathbf{28.53}$& \cellcolor{myblue}  $\mathbf{42.61}$ \\
					\hline
					\multirow{6}{*}{ \shortstack{Segformer$^{*}$}} & PGD&31.43&32.61&33.51&27.22&30.49&35.22&32.47&33.56&$\underline{1.85}$$^{*}$&43.16\\
					& SegPGD &31.85&32.98&34.47&27.79&31.11&36.09&32.86&34.05&3.20$^{*}$& 44.05\\
					& CosPGD &30.94&32.26&33.25&26.39&29.79&34.30&32.11&32.84&\cellcolor{myblue}  $\mathbf{1.70}$$^{*}$& 43.47\\
					& NI &\underline{23.68}&24.44&27.23&\underline{15.67}&20.90&26.59&\underline{24.01}&23.12&2.84$^{*}$&40.43 \\
					& MI &24.09&\underline{24.37}&\underline{26.93}&16.23&\underline{21.28}&\underline{26.21}&24.60&\underline{23.03}&2.09$^{*}$&\underline{38.95}\\
					
					&  EBAD &31.12&32.59&33.63&27.16&30.21&35.19&32.70&33.27&1.89$^{*}$&43.27\\
					&	   \textbf{Ours} & \cellcolor{mydblue}  $\mathbf{10.48}$ & \cellcolor{mydblue}  $\mathbf{13.08}$& \cellcolor{mydblue}  $\mathbf{14.36}$& \cellcolor{mydblue}  $\mathbf{8.63}$ & \cellcolor{mydblue}  $\mathbf{10.25}$ & \cellcolor{mydblue}  $\mathbf{14.10}$&\cellcolor{mydblue}   $\mathbf{12.78}$& \cellcolor{mydblue}  $\mathbf{11.11}$ &  $2.70$$^{*}$  &\cellcolor{myblue}   $\mathbf{37.52}$\\
					\bottomrule[1.2pt] 
				\end{tabular}
			}	
		\end{threeparttable}
	}
\end{table}
\begin{figure}[!t]
	\begin{center}
		\renewcommand\arraystretch{0.2}
		\begin{tabular}{@{\extracolsep{0.5em}}c@{\extracolsep{0.3em}}}
			\includegraphics[width=12.2cm,trim=0 0 0 0,clip]{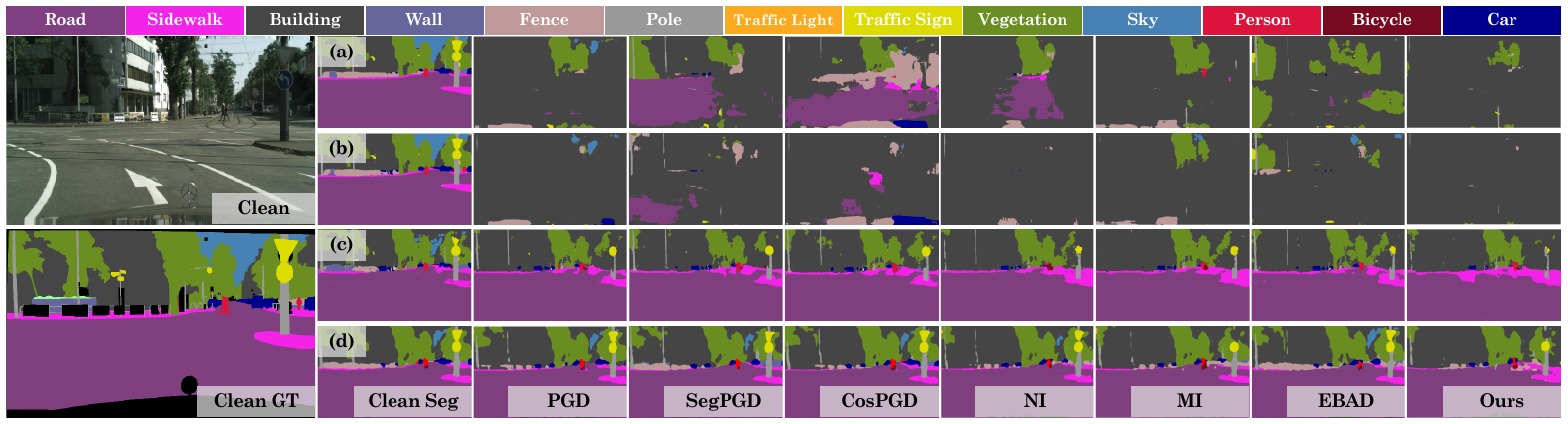}	\\
		\end{tabular}
	\end{center}
	\caption{Visualization of the attack results generated by 4 victim models on the Cityscapes dataset, including (a) DLV3-R101, (b) PSP-R101, (c) Segformer, and (d) Setr. We adopt Segformer as the surrogate model. }\label{fig:visual_1}
\end{figure}

\textbf{Qualitative Comparison.} In Fig.~\ref{fig:visual_1}, we present the attack results of $4$ segmentation models using DLV3-R50 as the surrogate model. As shown, all attack methods reduce visual quality, while \texttt{BMAT} produces the most degraded segmentation outputs, rendering CNN-based victims nearly ineffective. Additionally, for transformer victims, \texttt{BMAT} impairs the segmentation of key objects and regions.

\subsection{Mechanism Analysis}

\begin{figure}[!t]
	\centering
	\includegraphics[width=0.98\linewidth]{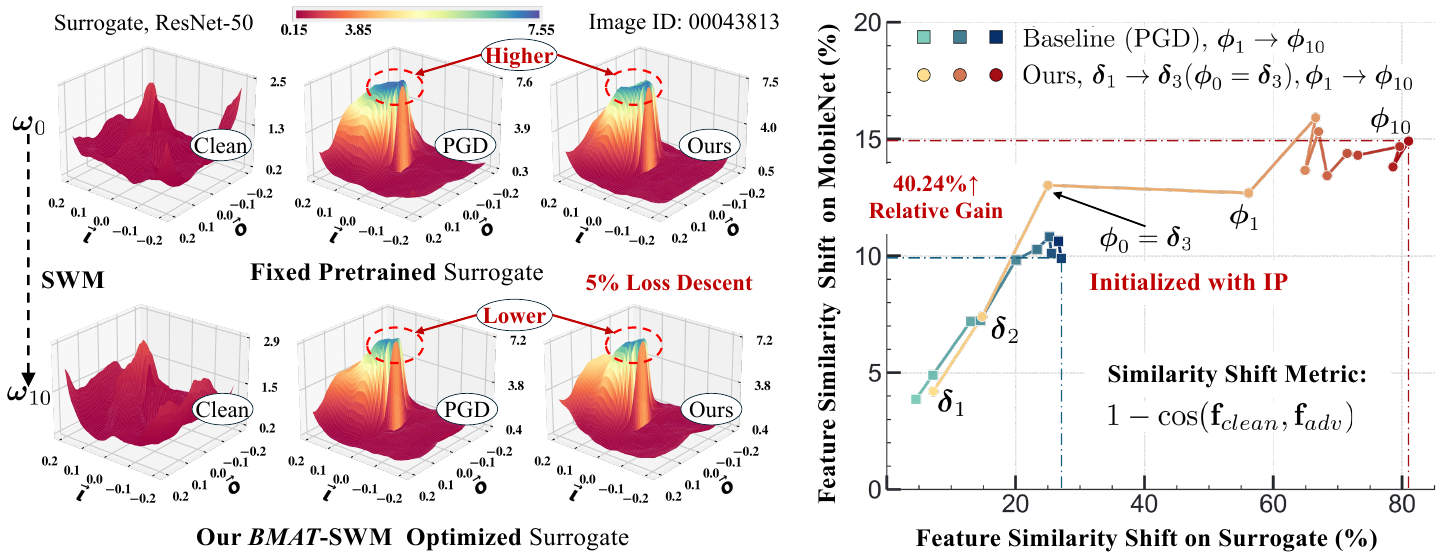}
	\caption{We visualize the tri-coupled coordination process of BMAT. \textbf{Left Panel:} It shows that SWM flattens the loss landscape of surrogate within $\sim$10 steps. \textbf{Right Panel:} It demonstrates that $\delta$ guides the trajectory toward a better feature-shift region. }\label{fig:onecol}
\end{figure}

\textbf{Tri-Coupled Coordination Dynamics.} 
We visualize the interaction among IP ($\delta$), perturbation ($\phi$), and surrogate adaptation ($\omega$) in Fig.~\ref{fig:onecol}. 
\emph{Left.} SWM reshapes the surrogate loss landscape within $\sim$10 steps, yielding $\sim$5\% loss descent and a smoother optimization region than the fixed pretrained surrogate. 
\emph{Right.} The learned IP evolves from noise ($\delta_1$) to structured patterns ($\delta_3$), steering the trajectory toward a higher feature shift, quantified as $1-\cos(f_{clean}, f_{adv})$, where $f(\cdot)$ denotes the normalized deep feature before the classification head. 
The induced perturbation initialized with $\phi_0=\delta_3$ produces a larger shift than PGD and yields a 40.24\% relative gain, indicating that transferability stems from coordinated variable evolution rather than isolated perturbation updates.

\begin{table}[!t]		
	\caption{Ablation analysis of  IGA  and SWM by employing MI as the base attacker. }
	\label{tab:atk_4}
	\centering
	\scalebox{0.7}{
		\begin{threeparttable} 
			\renewcommand\arraystretch{1.4}
			\setlength{\tabcolsep}{0.1mm}{
				\begin{tabular}{cccccccccccc}
					\toprule[1.2pt] 
					\multicolumn{12}{c}{ \normalsize Semantic Segmentation, Cityscapes dataset, mIoU $\downarrow$}\\
					\cmidrule{1-12}
					\multirow{2}{*}{\shortstack{Surrogate \\ Model}}&	\multirow{2}{*}{\shortstack{Basic \\ Attacker}}&\multicolumn{8}{c}{CNN }&\multicolumn{2}{c}{Transformer}\\
					\cmidrule(lr){3-10}\cmidrule(lr){11-12}  
					&  & FCN$^{\dag}$& UPerNet & PSANet & ISANet  &DLV3-R50$^{\ddag}$ &DLV3-R101&PSP-R50&PSP-R101 &Segformer& Setr   \\
					\hline \multirow{3}{*}{ FCN$^{\dag}$ }& MI &2.40$^{\dag}$&3.57&3.38&3.17&3.55&6.52&3.51&5.39&27.81& \underline{38.75}\\
					
					&{ MI+IGA}& \cellcolor{myblue} $\mathbf{1.56}$$^{\dag}$ &\cellcolor{myblue}  $\mathbf{2.40}$& \cellcolor{myblue} $\mathbf{2.20}$& \cellcolor{myblue} $\mathbf{1.98}$ & \cellcolor{myblue} $\mathbf{2.39}$ &\cellcolor{myblue}  $\mathbf{4.64}$&\cellcolor{myblue}  $\mathbf{2.32}$&\cellcolor{myblue}  $\mathbf{4.29}$ & \underline{${27.16}$}&${40.07}$\\
					&{ MI+IGA+SWM}& \underline{${1.75}$}$^{\dag}$& \underline{${2.74}$}& \underline{${2.60}$}& \underline{${2.42}$} & \underline{${2.81}$} & \underline{${5.30}$}& \underline{${2.62}$}& \underline{${4.44}$} & \cellcolor{myblue} $\mathbf{26.58}$&\cellcolor{myblue} $\mathbf{38.35}$\\ 
					\hline
					\multirow{3}{*}{ \shortstack{DLV3-R50$^{\ddag}$} } & MI &5.57&4.29&2.95&4.05&\cellcolor{myblue}  $\mathbf{0.99}$$^{\ddag}$&9.55&3.77&6.52&32.99&\underline{43.38}\\
					&{ MI+IGA}& \cellcolor{myblue}  $\mathbf{1.92}$ &\cellcolor{myblue}  $\mathbf{1.56}$& \cellcolor{myblue} $\mathbf{1.79}$& \cellcolor{myblue} $\mathbf{1.70}$ & \underline{${1.20}$}$^{\ddag}$ &\cellcolor{myblue}  $\mathbf{3.96}$&\cellcolor{myblue}  $\mathbf{1.48}$&\cellcolor{myblue}  $\mathbf{3.11}$ &  \underline{${29.78}$}& ${44.42}$\\
					&{ MI+IGA+SWM} & \underline{${2.60}$} & \underline{${1.96}$}& \underline{${2.16}$}& \underline{${2.31}$}& ${1.34}$$^{\ddag}$ & \underline{${4.54}$}& \underline{${1.86}$}& \underline{${3.57}$} & \cellcolor{myblue} $\mathbf{28.53}$& \cellcolor{myblue} $\mathbf{42.61}$ \\
					\bottomrule[1.2pt] 
				\end{tabular}
			}	
		\end{threeparttable}
	}
\end{table}

\textbf{IGA and SWM Modules.} As shown in Tab.~\ref{tab:atk_4}, IGA mainly strengthens \emph{within-architecture} transfer, consistently improving ASR on CNN victims. Adding SWM brings complementary gains on transformer victims, confirming that surrogate adaptation is especially useful for \emph{cross-architecture} transfer.

\textbf{Pseudo-surrogate Model.} Our main experiments use the auxiliary-surrogate setting, where $\mathcal{P}$ is instantiated by Inc-v3. To test the effectiveness of BMAT without extra victim access or architectural priors, we also instantiate $\mathcal{P}$ with a Bayesian version of the white-box surrogate via weight sampling~\cite{li2023making}. In this single-surrogate, zero-prior setting, \texttt{BMAT} still improves average ASR by 30.17\% and 58.34\% over the baselines (Tab.~\ref{tab:single_model}), showing that the gain mainly comes from its internal coordination. Using stronger CNN or transformer pseudo-surrogates further shifts transferability toward the corresponding victim families (Tab.~\ref{tab:atk_5}). This separation clarifies that the auxiliary pseudo-surrogate is not required for \texttt{BMAT}; it only provides an optional source of surrogate diversity, while the single-surrogate setting already verifies the bilevel trajectory effect.

\begin{table}[!t]		
	\centering
	\caption{ Ablation results of BMAT under the  single-surrogate zero-prior setting. }
	\label{tab:single_model}
	\scalebox{0.7}{
		\begin{threeparttable}
			\renewcommand\arraystretch{1.4}
			\setlength{\tabcolsep}{2.1mm}{
				\begin{tabular}{ccccccccccccc}
					\toprule[1.2pt]
					\multicolumn{12}{c}{\normalsize  Image Classification, {Ablation No additional victim model supervision}, ASR $\uparrow$} \\
					\cmidrule{1-12}
					\multicolumn{2}{c}{\multirow{2}{*}{Attacker}}&\multicolumn{4}{c}{CNN}&\multicolumn{3}{c}{CNN Ensemble}&\multicolumn{3}{c}{Transformer }\\
					\cmidrule(lr){3-6}\cmidrule(lr){7-9}\cmidrule(lr){10-12}
					&&  \scriptsize $[1]$   &\scriptsize $[2]$ &   \scriptsize$[3]$ & \scriptsize$[4]$&\scriptsize $[5]$& \scriptsize$[6]$ & \scriptsize$[7]$ &\scriptsize $[8]$&\scriptsize $[9]$&\scriptsize $[10]$   \\ \hline
					\multicolumn{2}{c}{ PGD}&  13.64 & 36.58 & 13.28 &  16.84 & 10.34 & 9.38 & 5.58 & 4.22 & 11.04 & 12.8 \\ 
					\multicolumn{2}{c}{ Ours  (Single-surrogate)}&  \cellcolor{myblue} {\bf 17.10}& \cellcolor{mydblue}  {\bf 63.16} & \cellcolor{myblue}  {\bf 17.04} &\cellcolor{mydblue}   {\bf 25.98} &\cellcolor{myblue}   {\bf 13.30} & \cellcolor{myblue}  {\bf 11.54} & \cellcolor{myblue}  {\bf 6.48} &\cellcolor{myblue}   {\bf 4.28} &\cellcolor{myblue}   {\bf 14.28} &\cellcolor{myblue}  {\bf 15.68}  \\ 
					\hline			                                                                                                                                                                                                                                                                                                                                                     
					\multicolumn{2}{c}{ PGD+MBA}& 15.34 & 57.22 & 14.04 & 20.40 & 11.10 & 10.26 & 6.26 & 3.78 & 12.22 & 13.28  \\ 
					\multicolumn{2}{c}{ Ours (Single-surrogate)}&\cellcolor{mydblue}   {\bf 26.08}& \cellcolor{mydblue}  {\bf 76.20} &  \cellcolor{mydblue} {\bf 25.56} &  \cellcolor{mydblue} {\bf 36.54} &\cellcolor{mydblue}   {\bf 19.30} &\cellcolor{mydblue}   {\bf 15.46} & \cellcolor{myblue}  {\bf 9.58} &  \cellcolor{myblue} {\bf 5.02} & \cellcolor{mydblue}  {\bf 19.98} &\cellcolor{myblue}  {\bf 19.28}  \\ 
					
					\bottomrule[1.2pt]
				\end{tabular}
			}
		\end{threeparttable}
	}
\end{table}

\begin{table*}[!t]		
	\caption{Ablation analysis by employing different pseudo-surrogate structures. $\dag$, $\ddag$, and $*$ denote the corresponding pseudo-surrogate. }\label{tab:atk_5}
	\centering
	\scalebox{0.7}{
		\begin{threeparttable} 
			\renewcommand\arraystretch{1.4}
			\setlength{\tabcolsep}{0.75mm}{
				\begin{tabular}{ccccccccccc}
					\toprule[1.2pt] 
					\multicolumn{11}{c}{ \normalsize  Semantic Segmentation, Cityscapes dataset, mIoU $\downarrow$}\\\cmidrule{1-11}
					\multirow{2}{*}{\shortstack{Surrogate \\ Model}}&	\multirow{2}{*}{\shortstack{Pseudo-surrogate \\Model}}&\multicolumn{7}{c}{CNN }&\multicolumn{2}{c}{Transformer}\\
					\cmidrule(lr){3-9}\cmidrule(lr){10-11} 
					&  & FCN$^{\dag}$& UPerNet & PSANet & ISANet &DLV3-R101&PSP-R50$^{\ddag}$&PSP-R101 &Segformer$^{*}$& Setr   \\
					\hline
					\multirow{3}{*}{ \shortstack{DLV3-R50} } & FCN$^{\dag}$&\underline{2.74}$^{\dag}$&\underline{2.11}&\underline{2.49}&2.60&\underline{4.42}&\underline{1.97}&\underline{3.49}&${29.92}$&45.50\\
					&\shortstack{PSP-R50$^{\ddag}$}& \cellcolor{myblue} $\mathbf{2.67}$ & \cellcolor{myblue} $\mathbf{1.76}$&\cellcolor{myblue}  $\mathbf{2.33}$&\cellcolor{myblue}  $\mathbf{2.19}$ &\cellcolor{myblue}  $\mathbf{2.85}$& \cellcolor{myblue} $\mathbf{1.77}^{\ddag}$ & \cellcolor{myblue} $\mathbf{2.68}$& ${31.01}$ & \underline{45.37}\\
					&\shortstack{Segformer$^{*}$}& ${3.51}$ &  ${2.85}$& ${2.55}$ & \underline{2.50}$^{\ddag}$ & ${4.76}$& ${2.21}$& ${4.09}$ &\cellcolor{mydblue}  $\mathbf{12.32}$$^{*}$& \cellcolor{myblue} $\mathbf{41.86}$ \\
					
					\bottomrule[1.2pt] 
				\end{tabular}
			}	
		\end{threeparttable}
	}
\end{table*}

\textbf{Analysis of Attack Iteration Fairness.}
Tab.~\ref{tab:iteration} reports a detailed ablation on the number of attack iterations.
PGD almost saturates around $K = 10$, and increasing $K$ to $20$ or $40$ brings only marginal or even degraded gains.
In contrast, \texttt{BMAT} with $(T,\tilde{K},K) = (2,5,10)$ and $(3,10,10)$ consistently outperforms PGD even when PGD is allowed to use more iterations; for example, at $K = 20$ and $40$, \texttt{BMAT} improves ASR by $+25.87\%$ and $+41.45\%$ over PGD on average. These results suggest that transferability is not improved by stronger surrogate over-optimization alone. Instead, \texttt{BMAT} improves trajectory-level generalization by separating the task-specific perturbation $\phi$ from the task-agnostic seed $\delta$.

\begin{table}[!t]
	\centering
	\caption{Ablation results of the attack iterations on the image classification tasks.}\label{tab:iteration}
	\scalebox{0.7}{
		\begin{threeparttable}
			\renewcommand\arraystretch{1.4}
			\setlength{\tabcolsep}{2.15mm}{
				\begin{tabular}{ccccccccccccc}
					\toprule[1.2pt]
					\multicolumn{12}{c}{\normalsize Image Classification, {Ablation of Attack Iterations}, ASR $\uparrow$ }\\
					\cline{1-12}
					\multicolumn{2}{c}{\multirow{2}{*}{Attacker}}&\multicolumn{4}{c}{CNN}&\multicolumn{3}{c}{CNN Ensemble}&\multicolumn{3}{c}{Transformer}\\
					\cmidrule(lr){3-6}\cmidrule(lr){7-9}\cmidrule(lr){10-12}
					&& \scriptsize $[1]$    &\scriptsize  $[2]$ &\scriptsize $[3]$  &\scriptsize $[4]$ &\scriptsize $[5]$ &\scriptsize $[6]$  &\scriptsize $[7]$&\scriptsize $[8]$&\scriptsize $[9]$ &\scriptsize $[10]$  \\ \hline
					\multicolumn{2}{c}{ PGD ,$K=10$}
					& 13.64 & 36.58 & 13.28 & 16.84 & 10.34 & 9.38 & 5.58 & 4.22 & 11.04 & 12.80\\ 
					\multicolumn{2}{c}{ PGD, $K=20$}& 13.92& 38.72 & 13.48 & 19.50& 10.10& 9.54 & 5.46& 4.48 & 12.56 & 13.70  \\ 
					\multicolumn{2}{c}{ Ours, $(T,\tilde{K},K)=(2,5,10)$}& {\underline{18.62}}& {\underline{ 43.40}}& {\underline{18.38}} & {\underline{22.04}} & {\underline{13.70}}& {\underline{13.02} }& {\underline{7.78}}& { \underline{5.82}} & {\underline{13.66} }& {\underline{15.10}}  \\ 
					\multicolumn{2}{c}{ PGD, $K=40$}& 13.22& 38.40 & 13.56 & 19.10& 9.32& 8.68 & 4.90& 4.28& 12.18 & 13.68  \\ 
					\multicolumn{2}{c}{ Ours, $(T,\tilde{K},K)=(3,10,10)$}& \cellcolor{myblue}{\bf  20.26}&\cellcolor{myblue} {\bf  47.86}& \cellcolor{myblue}{\bf 20.46} & \cellcolor{myblue}{\bf 24.72 }& \cellcolor{myblue}{\bf  14.58} &\cellcolor{myblue} {\bf 13.32 }& \cellcolor{myblue}{\bf 8.02}& \cellcolor{myblue}{\bf 6.04 }& \cellcolor{myblue}{\bf 15.18} & \cellcolor{myblue}{\bf 16.00}  \\ 
					\bottomrule[1.2pt]
				\end{tabular}
			}
		\end{threeparttable}
	}
\end{table}

\textbf{Fast and Full \texttt{BMAT}.}
To clarify the default implementation choice behind our main results, we compare the fast \texttt{BMAT} variant with a full variant in Tab.~\ref{tab:fast_full_bmat_main}. In the fast variant, Phase-I learns the trajectory seed through SWM-based perturbation-surrogate adaptation, and Phase-II uses standard sign/projection updates for efficient deployment. This also clarifies that Phase-I is not a plain warm-up: the SWM inner response directly shapes the IGA-based IP update, so the learned seed already encodes perturbation-surrogate joint adaptation before the fast Phase-II attack. The full variant retains SWM in Phase-II, leading to higher average ASR, especially on ensemble and transformer victims, but with additional runtime and memory under batch size $1$. These results show that the fast variant offers a practical accuracy-efficiency tradeoff, while the full variant further validates the benefit of maintaining joint adaptation throughout the attack trajectory.

\begin{table}[!t]
	\centering
	\caption{Fast/Full \texttt{BMAT} compared with strong combined baselines on ImageNet. Runtime and peak memory are measured with batch size $1$. ``Single-Surrogate'' uses the Bayesian version of the same ResNet-50 surrogate as $\mathcal{P}$; ``Auxiliary-Surrogate'' uses Inc-v3 as $\mathcal{P}$; ``Full'' retains SWM in Phase-II.}
	\label{tab:fast_full_bmat_main}
	\scalebox{0.7}{
		\begin{threeparttable}
			\renewcommand\arraystretch{1.4}
			\setlength{\tabcolsep}{2.5mm}
			\begin{tabular}{lccc|c|c|c}
				\toprule[1.2pt]
				\multicolumn{7}{c}{Image Classification, ResNet-50 surrogate, ASR $\uparrow$} \\
				\hline
				Method & CNN & CNN Ensemble & Transformer & Avg. ASR & Runtime & Peak Memory \\
				\hline
				DI-MI & 55.95 & 29.57 & 27.60 & 39.53 & \textbf{0.347}s & \textbf{399}MB \\
				\texttt{BMAT} (Fast, Single-Surrogate) & 66.75 & 37.97 & 28.13 & 46.53 & 1.489s & 1201MB \\
				\texttt{BMAT} (Full, Single-Surrogate) & \textbf{68.63} & 42.00 & 30.67 & 49.25 & 1.616s & 1290MB \\
				\texttt{BMAT} (Auxiliary-Surrogate) & 65.73 & \textbf{44.50} & \textbf{32.43} & \textbf{49.37} & 1.556s & 1157MB \\
				\hline
				DI-MI-MBA & 61.48 & 37.73 & 26.50 & 43.86 & \textbf{0.348}s & \textbf{992}MB \\
				\texttt{BMAT} (Fast, Single-Surrogate) & 65.23 & 42.13 & 26.77 & 46.76 & 1.509s & 1203MB \\
				\texttt{BMAT} (Full, Single-Surrogate) & \textbf{67.05} & 44.50 & \textbf{28.57} & \textbf{48.74} & 1.598s & 1295MB \\
				\texttt{BMAT} (Auxiliary-Surrogate) & 65.23 & \textbf{46.40} & 27.83 & 48.36 & 1.583s & 1153MB \\
				\bottomrule[1.2pt]
			\end{tabular}
		\end{threeparttable}
	}
\end{table}

\subsection{Ablation Study}
\begin{figure*}[!t]
	\begin{center}
		\renewcommand\arraystretch{0.2}
		\begin{tabular}{@{\extracolsep{0.1em}}c@{\extracolsep{-0.5em}}c@{\extracolsep{-0.5em}}c@{\extracolsep{0.5em}}}
			&&\\
			\includegraphics[width=4.2cm,trim=0 20 0 0,clip]{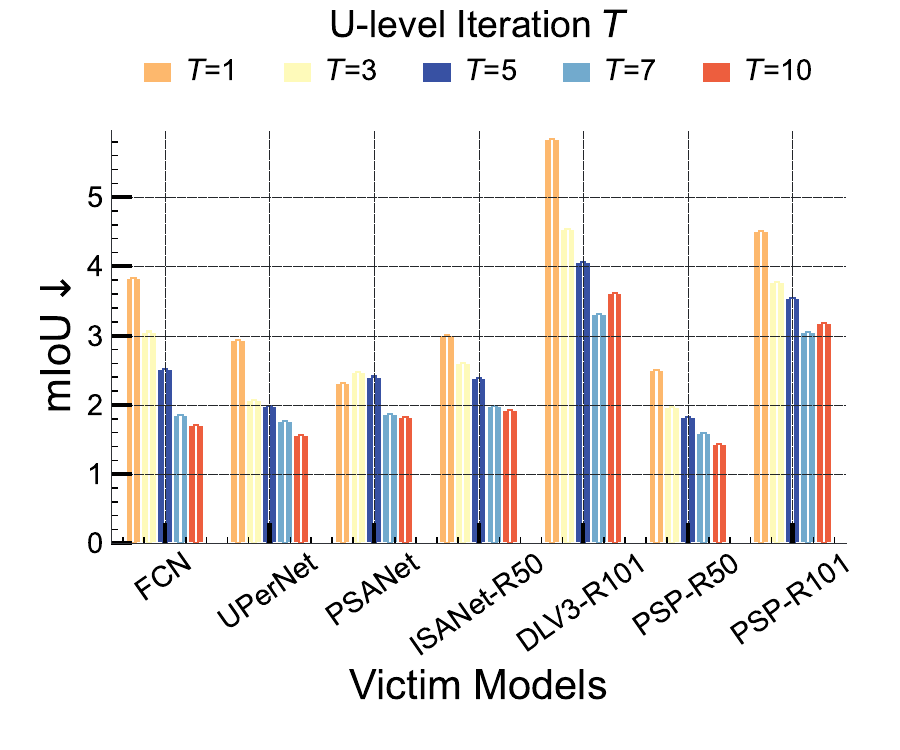}& \includegraphics[width=4.2cm,trim=0 20 0 0,clip]{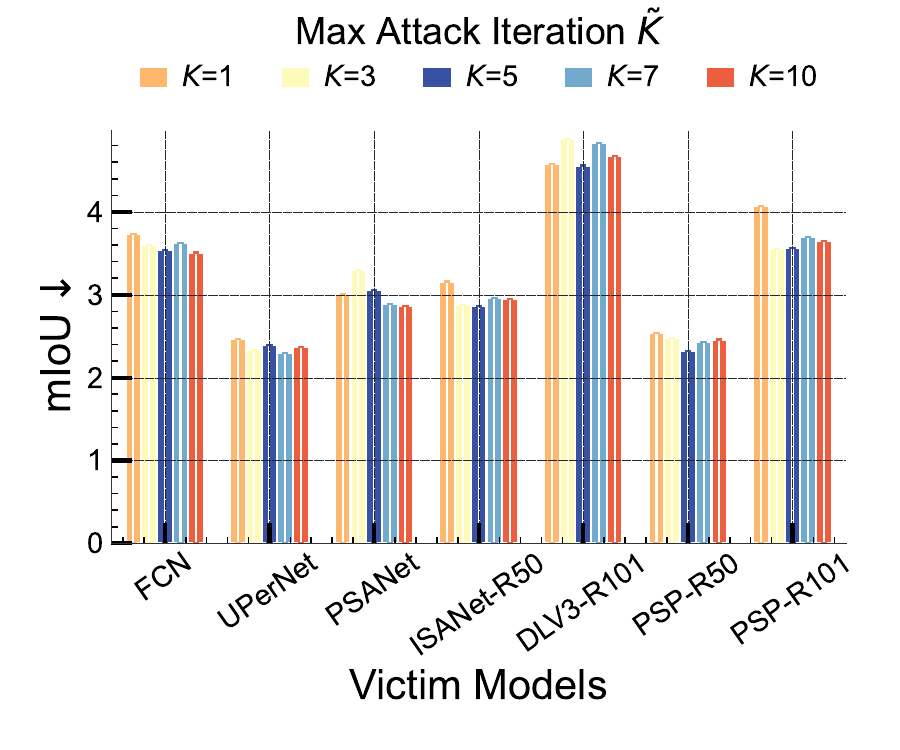}& \includegraphics[width=4.2cm,trim= 0 20 0 0,clip]{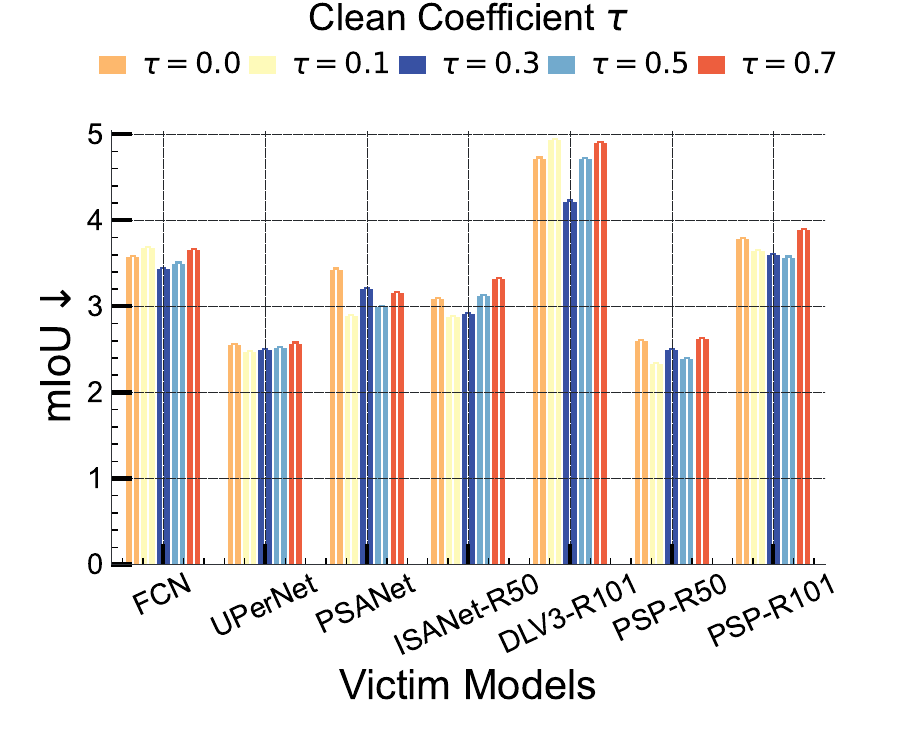}	\\
		\end{tabular}
	\end{center}
	\caption{Ablation results of 3 key hyperparameters across 7 CNN-based victim models on the semantic segmentation task. We adopt DLV3-R50 as the surrogate.}\label{fig:visual_2}
\end{figure*}

\textbf{Hyperparameters.} As shown in Fig.~\ref{fig:visual_2}, using larger $T$ leads to better transferability particularly from $1$ to $3$, while the larger $T$ also means more running cost. Using larger $\tilde{K}$ yields only marginal performance gains.  As for $\boldsymbol{\tau}$, relatively better performance is observed when ${\tau}=0.1$ or $0.5$.

\begin{wrapfigure}{r}{0.55\textwidth}
	\centering
	\includegraphics[width=0.55\textwidth,trim=0 10 0 0,clip]{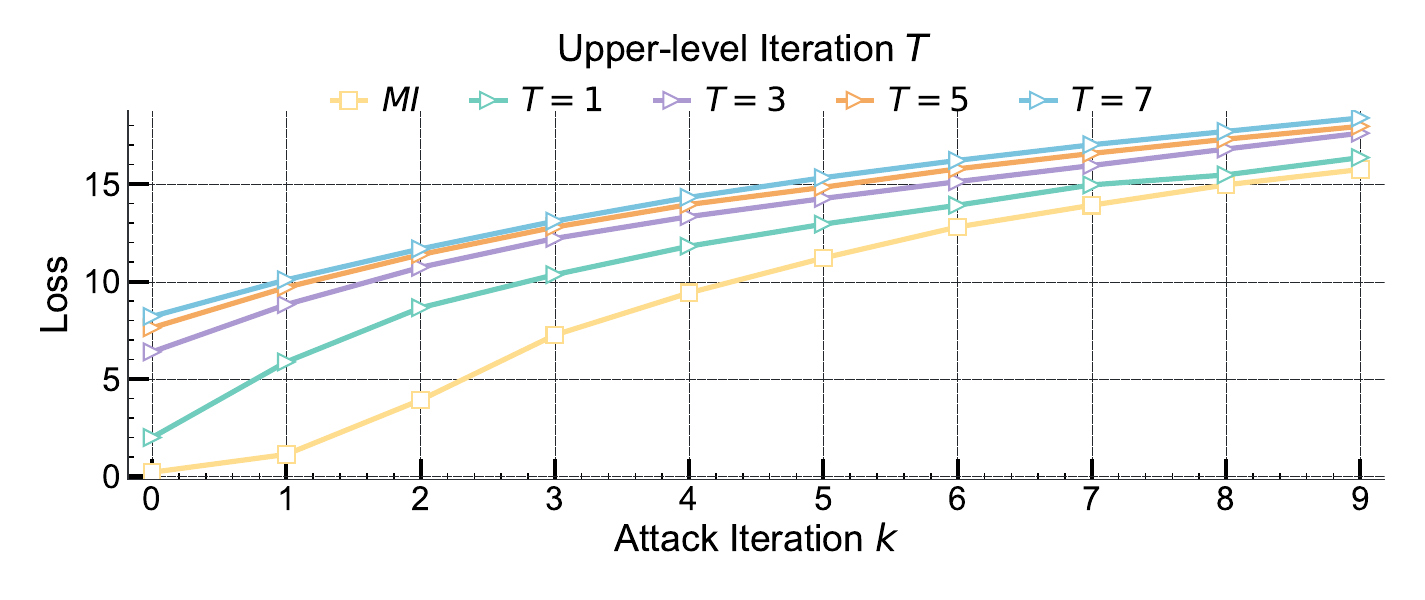}
	\caption{Illustrating the loss convergence behavior. We employ MI as the base attacker.}
	\label{fig:T_loss}
	\vspace{-2.2em}
\end{wrapfigure}

\textbf{Convergence Behavior.}
In Fig.~\ref{fig:T_loss}, we further analyze the optimization dynamics as $T$ and $K$ increase. Even a single IGA step ($T=1$) already leads to a clear loss decrease.
Increasing $T$ from $1$ to $3$ brings the most noticeable improvements, consistent with the hyperparameter trends as shown in Fig.~\ref{fig:visual_2}.

\begin{wraptable}{r}{0.55\textwidth}
	\vspace{-3.2em}
	\centering
	\caption{Runtime analysis as $T$ or $\tilde{K}$ increases. Note that VTA is implemented as PGD when $(T,\tilde{K})=0$.}
	\label{tab:runtime_analysis}
	\scalebox{0.7}{
		\begin{threeparttable}
			\renewcommand\arraystretch{1.4}
			\setlength{\tabcolsep}{1.5mm}{
				\begin{tabular}{ccccccc}
					\toprule[1.2pt]
					$\tilde{K}$ / $T$ & 0 (PGD) & 1 & 2 & 3 & 4 & 5 \\
					\cmidrule{1-7}
					0 (PGD) & 6.82 & N/A & N/A & N/A & N/A & N/A \\
					1 & N/A & 8.78 & 8.81 & 8.90 & 8.86 & 8.88 \\
					2 & N/A & 9.06 & 9.13 & 9.20 & 9.23 & 9.30 \\
					3 & N/A & 9.37 & 9.48 & 9.73 & 9.70 & 10.11 \\
					4 & N/A & 9.67 & 9.73 & 10.24 & 10.04 & 10.16 \\
					5 & N/A & 10.00 & 10.21 & 10.19 & 10.36 & 10.57 \\
					\bottomrule[1.2pt]
				\end{tabular}
			}
		\end{threeparttable}
	}
	\vspace{-1.5em}
\end{wraptable}

\textbf{Computational Efficiency.}
In Tab.~\ref{tab:runtime_analysis}, when $(T,\tilde{K})$ increase up to $3$, the average ASR gain over $10$ victims rises from $21.2\%$ to $21.54\%$ and $35.27\%$, while the average runtime only grows from $1.96$s to $2.31$s and $2.91$s.
Overall, the extra overhead introduced by \texttt{BMAT} optimization remains moderate and practically acceptable. We further verify in Tab.~\ref{tab:iteration} that simply increasing PGD iterations ($K=40$) leads to overfitting and performance degradation, whereas BMAT achieves a principled 41.45\% gain under a comparable budget.

\section{Conclusion}
We propose \texttt{BMAT}, a bilevel-minimax transfer attack framework that explicitly encodes the ternary coupling among IP, perturbation, and surrogate adaptation, and solves it via SWM and IGA with stability-oriented optimization dynamics. 
\textbf{Limitation.} \texttt{BMAT} incurs extra adaptation and hypergradient computation, suggesting future exploration of more lightweight first-order variants.

\bibliographystyle{splncs04}
    \bibliography{eccv26}
\end{document}